\documentclass[lettersize,journal]{IEEEtran}
\usepackage{amsmath,amsfonts}
\usepackage{algorithmic}
\usepackage{algorithm}
\usepackage{array}
\usepackage[caption=false,font=normalsize,labelfont=sf,textfont=sf]{subfig}
\usepackage{textcomp}
\usepackage{stfloats}
\usepackage{url}
\usepackage{verbatim}
\usepackage{graphicx}
\usepackage{booktabs}
\usepackage{cite}
\usepackage{makecell}
\newtheorem{definition}{Definition}
\newtheorem{proposition}{Proposition}
\usepackage{amsmath}
\usepackage{amssymb}

\usepackage{threeparttable}
\usepackage{siunitx}
\newtheorem{assumption}{Assumption}
\newtheorem{lemma}{Lemma}

\newenvironment{proof}
{\par\noindent\textit{Proof.}\ }
{\hfill\ensuremath{\square}\par}
\begin{document}

\title{ Reference-Model-Guided Distributed MPC for UAV Swarm Formation Flying Under Switching Directed Graphs }

\author{
	\vskip 1em

	Yiming WANG$^{1}$,
	Yenan WU$^{1}$,
    Junda WU$^{1}$,
    Yanke SUN$^1$,
    Huizhe LI$^1$, \\
    Ke YU$^2$,
    Youmin GONG$^1$,
    Guangfu MA$^1$,
    and Jie MEI$^1$.

	\thanks{
    Corresponding author: Jie Mei, \texttt{jmei@hit.edu.cn}.

    
    $^{1}$The authors are with the School of Intelligence
Science and Engineering, Harbin Institute of Technology (Shenzhen), Shenzhen 518055, China. 
		  
    $^{2}$The author is with the School of Electronic Engineering, Xidian University, Xi'an, 710071, China.
	}
}



\maketitle

\begin{abstract}
Distributed formation control of unmanned aerial vehicle (UAV) swarms in complex environments faces several challenges, including switching inter-agent connectivity, limited communication, external disturbances, and collision avoidance. To address these challenges, we propose a reference-model-guided distributed model predictive control (MPC) framework for UAV swarm formation. Each UAV uses relative-position information from its neighbors as input to a local reference model, whose output serves as the formation guidance for the onboard MPC controller. Meanwhile, a Kalman filter is utilized to predict neighboring UAV motions, enabling the construction of inter-UAV collision-avoidance constraints solely from instantaneous relative-state information. By decoupling MPC optimization from distributed coordination, this control architecture preserves the look-ahead planning and constraint-handling capabilities of MPC while enhancing the swarm's adaptability to switching directed graphs. Moreover, eliminating the exchange of neighbors' predicted trajectories reduces communication bandwidth demands, offering strong scalability for large-scale UAV swarms. To improve robustness and safety, disturbance-observer-based feedforward compensation is employed, while the bounded residual disturbance is explicitly considered in the control barrier function (CBF)-based obstacle-avoidance constraints. Finally, the proposed framework is validated through both simulations and real-world flight experiments. Specifically, the flight verification is conducted on a fully distributed multi-UAV platform without external global localization.
 \end{abstract}


\section{Introduction}
\IEEEPARstart{I}{n} recent years, with the rapid development of unmanned systems, unmanned aerial vehicle (UAV) swarms have been increasingly employed in applications such as cooperative patrol \cite{ma2024reinforcement, wang2021formation}, target tracking \cite{doostmohammadian2021distributed, zhao2024cooperative}, and collaborative payload transport \cite{wehbeh2020distributed, fink2011planning}. Formation control plays a fundamental role in enabling multiple UAVs to perform complex tasks collaboratively and has therefore attracted extensive research attention \cite{dong2018theory,yu2023adaptive,baldi2021adaptation,wang2023fully,wu2024distributed,shi2025synergistic,yan2021robust}. Compared with centralized architectures, distributed formation control relies primarily on information from neighboring UAVs, reducing dependence on a centralized controller and improving scalability for multi-UAV systems. However, practical deployment in complex environments remains challenging. Limited communication ranges, together with asymmetric or intermittent inter-UAV connectivity, can result in switching directed topologies, while external disturbances and collision-avoidance requirements must also be handled. This motivates the development of robust and scalable distributed formation-control methods for switching directed topologies.

\begin{figure}[htbp]
\centering
\includegraphics[width=1.0\linewidth]{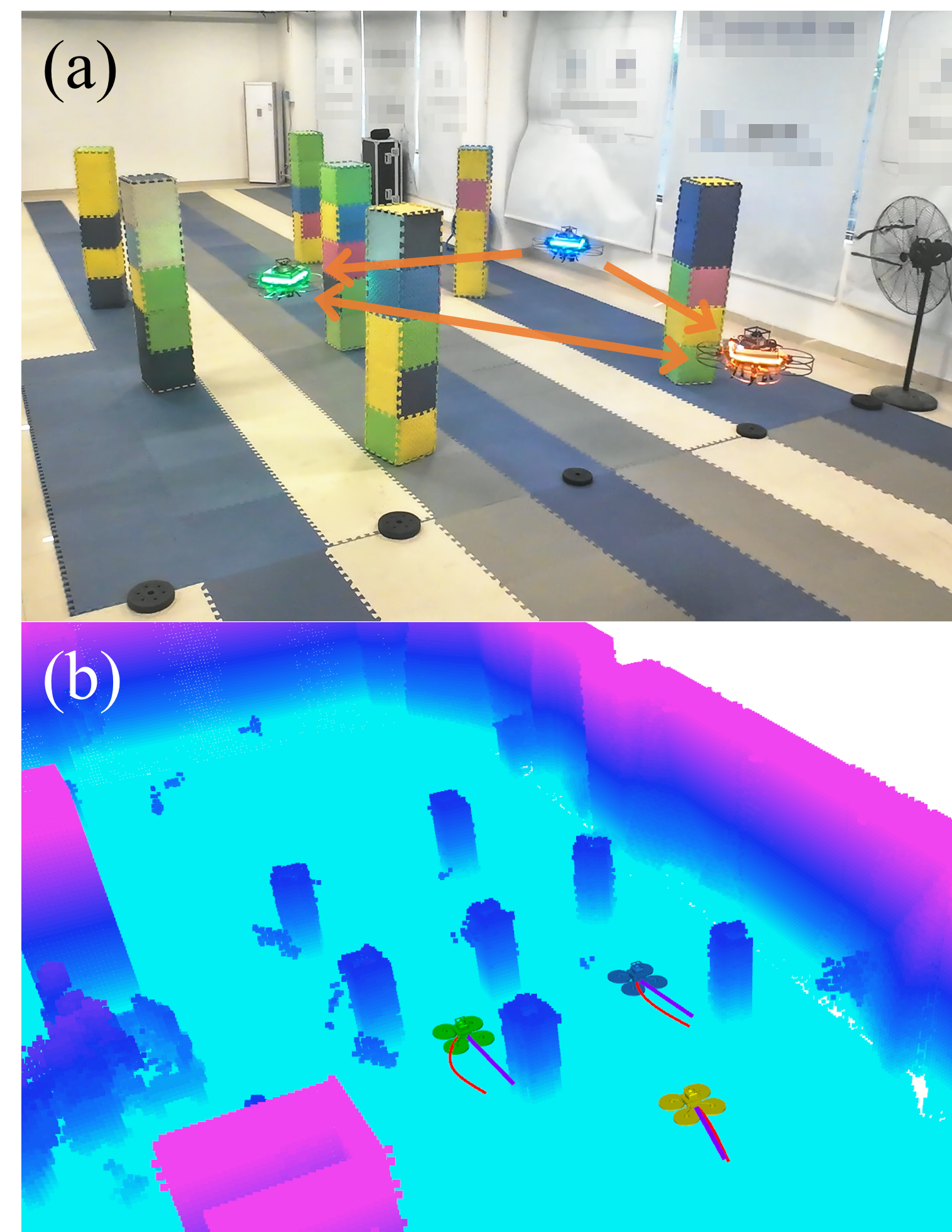}
\caption{Experimental verification of formation flight using three UAVs without external localization systems (e.g., GPS or motion capture). (a) The swarm traverses a dense obstacle region under wind disturbances. The orange arrows indicate the direction of information flow among the UAVs, where the communication topology of the formation at the current instant is a directed graph. (b) Visualization of online mapping and MPC outputs. The purple lines indicate the formation reference trajectories of the UAVs computed from relative states, while the red lines represent the predicted trajectories satisfying flight constraints including obstacle avoidance. }
\label{fig: first}
\end{figure}

For distributed UAV formation systems, the controller must achieve coordinated motion while satisfying safety and actuator constraints. Model predictive control (MPC) provides a natural framework for this purpose because it can explicitly incorporate system constraints and optimize predicted motion over a finite horizon. 
Accordingly, distributed MPC (DMPC) has been extensively
investigated for multi-agent formation control over
graph-based communication topologies
\cite{dunbar2006distributed,bono2021swarm,xiang2022energy}.
For multi-UAV systems, recent studies have further explored
DMPC-based formation control under directed communication
structures \cite{yuan2023distributed, chen2025distributed, hu2026fault}.
 However, many existing DMPC methods for cooperative multi-agent systems are formulated under predefined interaction structures \cite{zhou2022multiple, mi2019self, zheng2016distributed} and rely on the exchange of predicted or assumed state trajectories among neighboring agents for distributed coordination, including collision avoidance \cite{yuan2023distributed, katriniok2022fully, dunbar2011distributed}. Such horizon-wise information exchange inevitably increases the inter-UAV communication burden, since the transmitted data
volume grows with both the prediction horizon and the number of neighbors \cite{shorinwa2023distributed, izadi2011decentralized}, which is particularly undesirable in communication-limited networks with switching inter-UAV connectivity. These limitations motivate a distributed control architecture that can maintain formation coordination under switching directed topologies without requiring predicted-trajectory exchange.

In addition to communication and topology challenges, UAVs in practical flight inevitably encounter inter-UAV aerodynamic disturbances and external wind gusts. These disturbances degrade control accuracy and pose significant challenges to safe flight in dense-obstacle environments. To achieve disturbance rejection, existing studies have employed disturbance observers to estimate external disturbances and update the nominal model \cite{FixedTime}, while others have adopted disturbance-observer-based feedforward compensation to enable rapid adaptation with promising results \cite{lee2019robust}. However, such observers cannot comprehensively and accurately capture all disturbances, leaving non-negligible residual disturbances. Therefore, it remains challenging to guarantee safe flight in cluttered environments under residual disturbances.

To address these challenges, we propose a reference-model-guided distributed MPC framework for UAV swarm formation under switching directed topologies. For each UAV, a local reference model generates formation guidance using only instantaneous relative states. The onboard MPC controller then independently tracks this guidance while satisfying practical flight constraints. To avoid exchanging predicted trajectories for inter-UAV collision avoidance, each UAV locally predicts its neighbors' future motions from these relative states. For robust and safe flight, disturbance-observer-based feedforward compensation is employed to attenuate external disturbances, while the bounded residual disturbance is explicitly accounted for in the control barrier function (CBF)-based obstacle-avoidance constraints. Finally, the proposed framework is deployed and validated on a fully distributed multi-UAV platform without external localization systems, as shown in Fig.~\ref{fig: first}.

The main contributions are summarized as follows:
\begin{itemize}
\item \textit{Decoupling of Distributed Coordination and MPC Optimization}: A local reference model is introduced as an intermediate coordination layer within the distributed MPC framework, generating formation guidance solely from instantaneous relative states. This architecture simplifies the onboard optimization formulation while adapting effectively to switching directed communication topologies. Meanwhile, it preserves the look-ahead planning and constraint-handling capabilities of MPC.

\item \textit{Instantaneous-State-Only Interaction without Trajectory Exchange}: To preserve the trajectory-exchange-free property during inter-UAV avoidance, each UAV locally predicts its neighbors' motions solely from instantaneous relative states. The predicted motions are then incorporated into the inter-UAV collision-avoidance constraints. This design reduces communication bandwidth overhead and facilitates scaling to large swarms.

\item \textit{Disturbance Compensation and Safe Obstacle Avoidance}: Disturbance-observer-based feedforward compensation is integrated into the local controller to actively reject external disturbances. Meanwhile, the bounded residual disturbance is explicitly incorporated into the CBF-based obstacle-avoidance constraints. This formulation improves flight robustness and enhances safety against wind gusts in cluttered environments.

\item \textit{Fully Distributed Validation without External Localization}: The proposed framework is implemented on a fully distributed multi-UAV platform without external localization systems (e.g., GPS or motion capture). Each UAV independently performs onboard state estimation, neighbor-state prediction, and local MPC optimization. Extensive simulations and real-world flight experiments validate the effectiveness and practicality of the overall system.
\end{itemize}

\section{Preliminaries and Problem Formulation}

\subsection{UAV Dynamics}\label{subsec:quad_model}

Consider a formation system consisting of $n$ UAVs. For the $i$th UAV, let $\boldsymbol{p}^i\in\mathbb{R}^{3}$ and $\boldsymbol{v}^i\in\mathbb{R}^{3}$ denote its position and velocity in the inertial frame, respectively, and let $\boldsymbol{q}^i\in\mathbb{R}^{4}$ denote its unit-quaternion attitude. The low-level control inputs consist of the collective thrust command $T_{\mathrm{cmd}}^i\in\mathbb{R}$ and the body angular velocity command $\boldsymbol{\omega}_{\mathrm{cmd}}^i\in\mathbb{R}^{3}$, which are subsequently tracked by the PX4 rate controller \cite{meier2015px4}.

Following \cite{faessler2017differential, faessler2016thrust}, the nonlinear dynamics of the $i$th UAV are given by
\begin{subequations}\label{eqn: non_lin_dynamics}
    \begin{align}
        \dot{\boldsymbol{p}}^i
        &= \boldsymbol{v}^i,\\
        \dot{\boldsymbol{v}}^i
        &= \frac{1}{m^i}\boldsymbol{R}(\boldsymbol{q}^i)\boldsymbol{T}^i
        - \boldsymbol{g}
        + \boldsymbol{d}^i,
        \label{eq:vel_with_dis}\\
        \dot{\boldsymbol{q}}^i
        &= \frac{1}{2}\boldsymbol{q}^i \otimes
        \begin{bmatrix}
            0\\
            \boldsymbol{\omega}_{\text{cmd}}^i
        \end{bmatrix},
    \end{align}
\end{subequations}
where $m^i$ denotes the mass of the $i$th UAV,
$\boldsymbol{R}(\boldsymbol{q}^i)\in\mathbb{R}^{3\times3}$ is the
rotation matrix from the body frame to the inertial frame corresponding to $\boldsymbol{q}^i$,
$\boldsymbol{T}^i=[0,0,T_{\text{cmd}}^i]^{\mathrm{T}}$ is the thrust vector expressed in
the body frame, $\boldsymbol{g}=[0,0,g]^{\mathrm{T}}$ denotes the
gravitational acceleration, and the operator $\otimes$ denotes quaternion
multiplication.
The term $\boldsymbol{d}^i\in\mathbb{R}^{3}$ represents the lumped
acceleration disturbance, including wind-induced disturbances and
unmodeled translational dynamics.

In practical UAV systems, a cascaded control architecture is widely adopted \cite{faessler2017differential, liu2024ipc}, where an outer-loop controller regulates translational motion and an inner-loop controller tracks attitude commands. When the inner-loop attitude controller operates at a significantly higher bandwidth than the outer-loop translational controller, the attitude response can be assumed to be sufficiently fast relative to the translational dynamics. Under this assumption, the UAV translational dynamics can be approximated by
\begin{subequations}
    \begin{align}
        \dot{\boldsymbol{p}}^i
        &= \boldsymbol{v}^i,\\
        \dot{\boldsymbol{v}}^i
        &= \boldsymbol{u}^i
        - \boldsymbol{g}
        + \boldsymbol{d}^i,
    \end{align}
    \label{eqn:trans_dynamics}
\end{subequations}
where $\boldsymbol{u}^i = \frac{1}{m^i}\boldsymbol{R}(\boldsymbol{q}^i)\boldsymbol{T}^i \in \mathbb{R}^3$ denotes the virtual acceleration-level control input. In our framework, $\boldsymbol{u}^i$ is computed by the outer-loop MPC with disturbance compensation. It is then mapped to the low-level commands $(T_{\mathrm{cmd}}^i, \boldsymbol{\omega}_{\mathrm{cmd}}^i)$ via differential flatness and a nonlinear attitude controller \cite{faessler2017differential}.

\subsection{Switching Directed Interaction Graphs}
\label{subsec:graph_theory}

The interaction topology of a formation consisting of $n$ UAVs is modeled by a switching directed graph $\mathcal{G}[k]=(\mathcal{V},\mathcal{E}[k])$, where $k$ is the discrete-time index, $\mathcal{V}=\{\xi_1,\xi_2,\ldots,\xi_n\}$ denotes the vertex set and $\mathcal{E}[k]\subseteq\mathcal{V}\times\mathcal{V}$ denotes the time-varying edge set. An edge $(\xi_j,\xi_i)\in\mathcal{E}[k]$ indicates that UAV $i$ can receive information from UAV $j$. Accordingly, the in-neighbor set of UAV $i$ is defined as $\mathcal{N}_i[k]=\{\xi_j\in\mathcal{V}:(\xi_j,\xi_i)\in\mathcal{E}[k]\}$.

A directed path from $\xi_i$ to $\xi_j$ is a sequence of directed edges $(\xi_i,\xi_{i_2}),(\xi_{i_2},\xi_{i_3}),\ldots,(\xi_{i_k},\xi_j)$. A directed graph is said to have a directed spanning tree if there exists a node, called the root, from which every other node is reachable through a directed path.

The adjacency matrix associated with $\mathcal{G}[k]$ is denoted by $\mathcal{A}[k]=[a_{ij}[k]]\in\mathbb{R}^{n\times n}$, where $a_{ij}[k]=1$ if $(\xi_j,\xi_i)\in\mathcal{E}[k]$ and $a_{ij}[k]=0$ otherwise. Self-edges are excluded, and hence $a_{ii}[k]=0$. The corresponding degree matrix is defined as $\mathcal{D}[k]=\operatorname{diag}(d_1[k],\ldots,d_n[k])$, where $d_i[k]=\sum_{j=1,j\neq i}^{n}a_{ij}[k]$. The Laplacian matrix of $\mathcal{G}[k]$ is defined as $\mathcal{L}[k]=\mathcal{D}[k]-\mathcal{A}[k]=[l_{ij}[k]]\in\mathbb{R}^{n\times n}$, where $l_{ii}[k]=d_i[k]$ and $l_{ij}[k]=-a_{ij}[k]$ for $i\neq j$.

\begin{definition}
A matrix $M\in\mathbb{R}^{n\times n}$ is nonnegative if all its entries are nonnegative.
\end{definition}

\begin{definition}
A nonnegative matrix $M\in\mathbb{R}^{n\times n}$ is row-stochastic if all its row sums are $1$.
\end{definition}

\begin{definition}
A row-stochastic matrix $M\in\mathbb{R}^{n\times n}$ is called ergodic if there exists a nonnegative vector $\boldsymbol{c}\in\mathbb{R}^{n}$ such that $\lim_{k\rightarrow\infty}M^{k}=\mathbf{1}_n\boldsymbol{c}^{T}$, where $\mathbf{1}_n\in\mathbb{R}^n$ denotes the column vector of all ones.
\label{def:ergodic}
\end{definition}

To characterize the switching communication topology, we adopt the following assumptions commonly used in the analysis of consensus and coordination over switching directed graphs \cite{jadbabaie2003coordination, ren2005consensus, qin2012consensus}.
\begin{assumption}
All possible communication graphs $\mathcal{G}[k]$ belong to a finite set $\bar{\mathcal{G}} = \{\mathcal{G}_1,\mathcal{G}_2,\ldots,\mathcal{G}_{N_g}\}$, where $N_g$ denotes the number of admissible communication topologies.
\label{assump:finiteset}
\end{assumption}

\begin{assumption}
There exists an infinite sequence of contiguous and nonempty time intervals $[k_r,k_{r+1})$, $r=1,2,\ldots$, with $k_1=0$ and $1 \le k_{r+1}-k_r \le \sigma$ for some positive integer $\sigma$, such that the union of the directed graphs $\mathcal{G}[k_r],\mathcal{G}[k_r+1],\ldots,\mathcal{G}[k_{r+1}-1]$ over each interval $[k_r,k_{r+1})$ contains a directed spanning tree.
\label{assump:spanningtree}
\end{assumption}

\subsection{Discrete-Time Control Barrier Functions}

Control barrier functions (CBFs) provide a systematic approach to enforcing safety constraints in dynamical systems. Given a continuously differentiable barrier function $b(\boldsymbol{x})$, the associated safe set is defined as
\begin{equation}
    \mathcal{X}_{\mathrm{safe}}=\{\boldsymbol{x}\in\mathbb{R}^{n_x}\mid b(\boldsymbol{x})\geq 0\},
\end{equation}
where $n_x$ denotes the state dimension.
For a discrete-time system, the forward invariance of the safe set can be ensured by imposing the discrete-time exponential CBF condition \cite{10807140}:
\begin{equation}
    b(\boldsymbol{x}[k+1])\geq(1-\lambda_o)b(\boldsymbol{x}[k]),
    \label{equ:cbf_def}
\end{equation}
where $\lambda_o\in(0,1]$ is a design parameter that determines the admissible decrease rate of the barrier function. This condition guarantees the forward invariance of the safe set, keeping the barrier function nonnegative for any initial state in $\mathcal{X}_{\mathrm{safe}}$.

\subsection{Problem Formulation}
\label{subsec:problem_formulation}
Consider a swarm of $n$ UAVs interacting over switching directed graphs $\mathcal{G}[k]$. Design a distributed formation controller such that the prescribed formation is achieved and the desired common velocity $\boldsymbol{v}^{*}\in\mathbb{R}^{3}$ is tracked, while satisfying the input and collision-avoidance constraints and effectively handling environmental disturbances.

Let $\boldsymbol{\delta}^i\in\mathbb{R}^{3}$ denote the prescribed formation offset of UAV $i$, and define the desired relative displacement between UAVs $i$ and $j$ as $\boldsymbol{\delta}^{ij}=\boldsymbol{\delta}^i-\boldsymbol{\delta}^j$. The formation-control objective is to drive the UAV swarm toward the prescribed configuration while tracking the desired common velocity, such that the formation errors are uniformly ultimately bounded:
\begin{equation}
\begin{aligned}
&\limsup_{k\rightarrow\infty}
    \left\|
    \boldsymbol{p}^i[k]-\boldsymbol{p}^j[k]-\boldsymbol{\delta}^{ij}
    \right\|_2
    \leq \epsilon_p,\\
    &\limsup_{k\rightarrow\infty}
    \left\|
    \boldsymbol{v}^i[k]-\boldsymbol{v}^{*}
    \right\|_2
    \leq \epsilon_v,
\end{aligned}
\end{equation}
for all $i,j\in\{1,\ldots,n\}$, where $\epsilon_p, \epsilon_v > 0$ denote the ultimate bounds of the position and velocity formation errors, respectively.

\begin{figure*}[t]
\centering
\includegraphics[width=\textwidth]{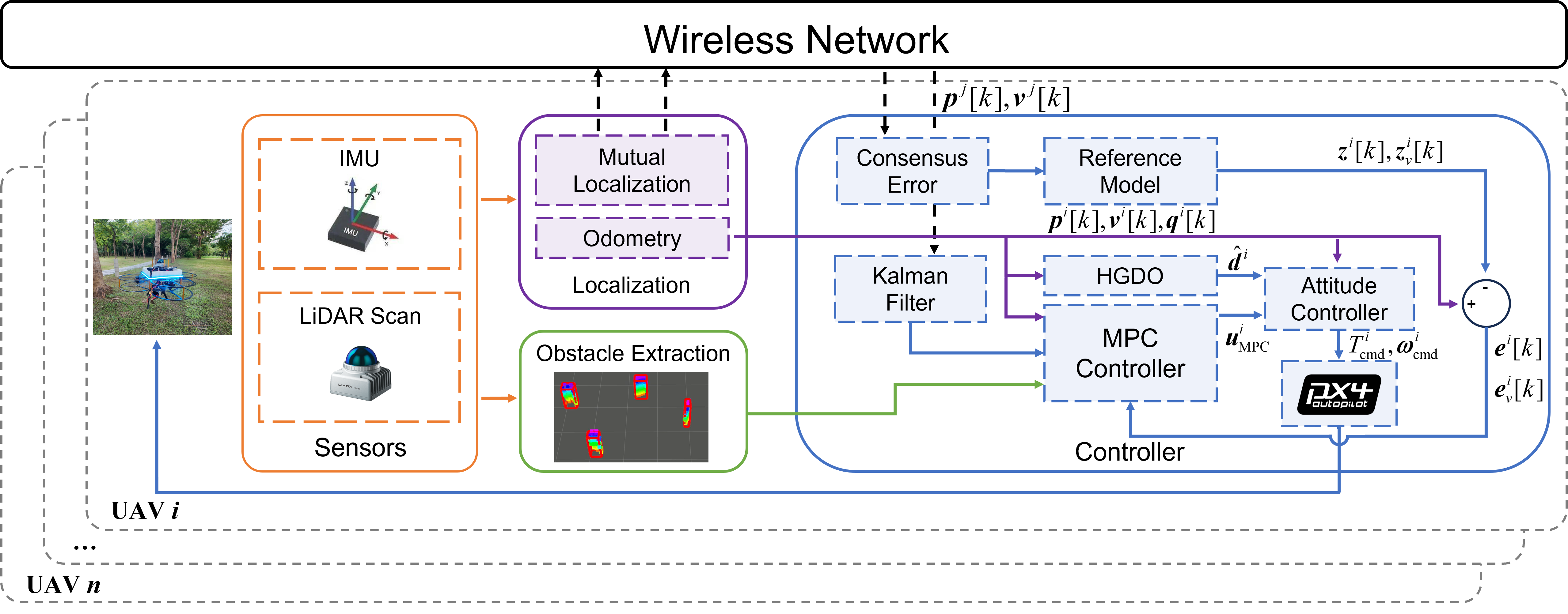}
\caption{System overview of the proposed distributed framework. On each UAV, Swarm-LIO2 \cite{zhu2024swarm} fuses onboard LiDAR, inertial measurement unit (IMU), mutual teammate observations, and wireless network data to estimate the ego state ($\boldsymbol{p}^i[k], \boldsymbol{v}^i[k], \boldsymbol{q}^i[k]$) and neighboring states ($\boldsymbol{p}^j[k], \boldsymbol{v}^j[k]$). Meanwhile, clustered LiDAR point clouds are used for obstacle extraction to construct CBF-based obstacle-avoidance constraints. Neighboring relative states drive the local reference model to generate the reference position $\boldsymbol{z}^i[k]$ and velocity $\boldsymbol{z}_v^i[k]$, yielding the tracking errors $\boldsymbol{e}^i[k]$ and $\boldsymbol{e}_v^i[k]$ for the local MPC. In addition, a Kalman filter predicts neighboring UAV motions for priority-based inter-UAV collision-avoidance constraints. Finally, a high-gain disturbance observer (HGDO) estimates the lumped translational disturbance $\hat{\boldsymbol{d}}^i$ for feedforward compensation, and the attitude controller converts the compensated acceleration command $\boldsymbol{u}_{\mathrm{MPC}}^i - \hat{\boldsymbol{d}}^i$ into $T_{\mathrm{cmd}}^i$ and $\boldsymbol{\omega}_{\mathrm{cmd}}^i$ for the PX4 rate controller.}
\label{fig: system overview}
\end{figure*}

\section{Methodology}
\subsection{System Overview}
In this paper, we propose a reference-model-guided distributed MPC framework for autonomous UAV swarm formation. A local reference model is introduced as an intermediate layer that decouples distributed formation coordination from local constrained optimization. Specifically, instantaneous relative-state information from neighboring UAVs drives the reference model, whose output provides formation guidance for the local MPC. Accordingly, the control design is organized into two interconnected components: 1) distributed formation coordination based on the reference model, and 2) local MPC for reference tracking and constraint handling, integrated with disturbance feedforward compensation.

The overall architecture of the proposed method is illustrated in Fig.~\ref{fig: system overview}. While the physical UAV motion follows the nonlinear dynamics in \eqref{eqn: non_lin_dynamics}, the local MPC employs the reduced-order linear translational model \eqref{eqn:trans_dynamics} for motion prediction and constrained optimization. Within this cascaded flight-control architecture, an inner-loop nonlinear attitude controller converts the outer-loop MPC acceleration command into collective-thrust and body-rate commands, which are then tracked by the PX4 rate controller.

To reject environmental disturbances, a high-gain disturbance observer (HGDO) estimates the lumped translational disturbance and applies feedforward compensation at the acceleration-command level. Environmental obstacle information is extracted from onboard LiDAR measurements and used to construct CBF-based obstacle-avoidance constraints, in which the residual disturbance is explicitly accounted for to enhance safety robustness.  Meanwhile, each UAV receives the instantaneous positions and velocities of its neighbors over the wireless network and locally predicts their future motions using a Kalman filter. These local predictions are then incorporated into the inter-UAV collision-avoidance constraints, eliminating the need to exchange predicted trajectories over the MPC horizon.

\par

\subsection{Reference Model Design}
Based on neighboring information, each UAV independently generates a local formation reference through the reference model, which provides guidance for the local MPC.

We design a distributed control input for the reference model,
  which is modeled as the following discrete-time double-integrator system:

   \begin{flalign}
	\begin{split}
	\boldsymbol{z}^{i}[k+1]&=\boldsymbol{z}^{i}[k]+h\boldsymbol{z}_v^i[k],\\
	\boldsymbol{z}_v^i[k+1]&=\boldsymbol{z}_v^i[k]+h\boldsymbol{u}^i_{f}[k],
	\label{referencemodel_closeloop}
   \end{split}
   \end{flalign}
   where $h$ denotes the discretization step size of the reference and prediction models. The discrete-time index is denoted by $k$, $\boldsymbol{z}^{i}[k]\in\mathbb{R}^3$ and $\boldsymbol{z}_v^{i}[k]\in\mathbb{R}^3$ are the reference model outputs, and $\boldsymbol{u}^i_{f}[k]\in\mathbb{R}^3$ is the control input of the reference model.
   Inspired by the distributed consensus algorithm for discrete-time second-order dynamics over switching directed graphs \cite{qin2012consensus}, we design the following reference-model control input for UAV $i$:
\begin{flalign}
 \begin{split}
 \boldsymbol{u}^i_{f}[k]=-\sum_{j=1}^n a_{i j}[k]\left(\boldsymbol{p}^i[k]-\boldsymbol{p}^j[k]-\boldsymbol{\delta}^{ij}\right)
 -2{\gamma}\left(\boldsymbol{v}^i[k]-\boldsymbol{v}^*\right).
 \label{formation_second_order}
\end{split}
\end{flalign}
In the above, $\boldsymbol{p}^i[k]$ and $\boldsymbol{v}^i[k]$ are the practical position and velocity of UAV $i$, $\gamma\in \mathbb{R}^+$ is velocity damping gain, $a_{ij}[k]$ are the entries of the adjacency matrix $\mathcal{A}[k]$, $i,j=1,\ldots,n$, $\boldsymbol{\delta}^{ij}$ is the desired relative displacement, and
$\boldsymbol{v}^*\in \mathbb{R}^3$ is the prescribed formation maneuvering velocity.

\subsection{Model Predictive Control Algorithm Design}\label{subsec:MPC}

Define the tracking errors between the actual UAV states and the corresponding reference-model outputs as
\begin{equation}
	 \boldsymbol{e}^i[k]=\boldsymbol{p}^i[k]-\boldsymbol{z}^i[k],\quad \boldsymbol{e}_v^i[k]=\boldsymbol{v}^i[k]-\boldsymbol{z}_v^i[k].
	 \label{trackingerr}
\end{equation}

To achieve the formation objective, each UAV is required to track the position and velocity references generated by its local reference model, such that the tracking errors $\boldsymbol{e}^i[k]$ and $\boldsymbol{e}_v^i[k]$ are uniformly ultimately bounded (UUB).
Accordingly, an MPC controller is developed for each UAV to track the
reference-model outputs.
Directly incorporating the nonlinear UAV dynamics \eqref{eqn: non_lin_dynamics} into the MPC formulation can significantly increase the computational burden \cite{sun2022comparative}, particularly when obstacle-avoidance and inter-UAV collision-avoidance constraints are considered. 
Taking advantage of the cascaded flight-control architecture, the UAV translational dynamics are approximated by a discrete-time double-integrator model for MPC prediction:
\begin{equation}
\begin{bmatrix}
\boldsymbol{p}[k+1] \\
\boldsymbol{v}[k+1]
\end{bmatrix}
=
\begin{bmatrix}
\boldsymbol{I}_3 & h \boldsymbol{I}_3 \\
0 & \boldsymbol{I}_3
\end{bmatrix}
\begin{bmatrix}
\boldsymbol{p}[k] \\
\boldsymbol{v}[k]
\end{bmatrix}
+
\begin{bmatrix}
\frac{1}{2} h^2 \boldsymbol{I}_3 \\
h \boldsymbol{I}_3
\end{bmatrix}
\boldsymbol{u}[k],
\label{equ:discrete_double_integrator}
\end{equation}
where $\boldsymbol{p}$, $\boldsymbol{v}$, and $\boldsymbol{u}$ denote the
position, velocity, and acceleration control input, respectively.
$\boldsymbol{I}_3\in\mathbb{R}^{3\times3}$ represents the identity matrix,
and $h$ denotes the prediction discretization interval. 
 In the following, we consider $ N \in \mathbb{N}^+$ as the prediction horizon of MPC. The notation $\hat{(\cdot)}\left[l \mid k\right]$ represents the predicted value at $k+l$ with the information acquired at $k$, where $l\in\left\{0,\dots,N-1\right\}$.

Although the double-integrator model provides an efficient prediction
model for online optimization, external disturbances such as wind
perturbations may degrade the tracking performance of the UAVs.
Therefore, the lumped translational disturbance is estimated and
compensated through a high-gain disturbance observer (HGDO) based on the
dynamics model \eqref{eq:vel_with_dis}:
\begin{subequations}
\begin{align}
    \dot{\hat{\boldsymbol{v}}}^i &= \frac1{m^i} \boldsymbol{R}(\boldsymbol{q}^i) \boldsymbol{T}^i - \boldsymbol{g} + \hat{\boldsymbol{d}}^i
                        + \frac{\alpha_{1}}{\alpha_3}\bigl(\boldsymbol{v}^i-\hat{\boldsymbol{v}}^i\bigr), \\
    \dot{\hat{\boldsymbol{d}}}^i &= \frac{\alpha_{2}}{\alpha_3^{2}}\bigl(\boldsymbol{v}^i-\hat{\boldsymbol{v}}^i\bigr),
\end{align}
\label{equ:HGDO}
\end{subequations}
\noindent where $\alpha_1,\alpha_2$  are observer gains, $\alpha_3$ is a small tuning parameter, $\hat{\boldsymbol{v}}^i$ and $\hat{\boldsymbol{d}}^i$ denote the estimation of $\boldsymbol{v}^i$ and $\boldsymbol{d}^i$, respectively. 
The estimated disturbance is compensated through a feedforward term
combined with the MPC-generated acceleration command.

 Since the reference model outputs may entail collision risks, obstacle-avoidance
constraints and inter-UAV collision-avoidance constraints are further
incorporated into the MPC formulation.

\par
First, we formulate the MPC cost function 
to penalize the reference-tracking error
and the control effort over the prediction horizon.
The weighted quadratic norm is defined as
$\|\boldsymbol{x}\|_{\boldsymbol{W}}^2
=\boldsymbol{x}^{\mathrm{T}}\boldsymbol{W}\boldsymbol{x}$,
where $\boldsymbol{W}$ denotes the corresponding weighting matrix.

$1)$ \emph{Tracking Error Penalty}:
\begin{equation}
 \ell_{e}^i=\left\|\hat{\boldsymbol{x}}^i[l|k]-\boldsymbol{x}_{r}^i[l|k]\right\|^2_{\boldsymbol{W}^i_e},
\end{equation}
where $\boldsymbol{W}^i_e\in\mathbb{R}^{6 \times 6}$ is the weighting matrix, 
$\hat{\boldsymbol{x}}^i[l|k]=[\hat{\boldsymbol{p}}^i[l|k], \hat{\boldsymbol{v}}^i[l|k]]^{\mathrm{T}}\in\mathbb{R}^6$ and $\boldsymbol{x}_{r}^i[l|k]=[\boldsymbol{z}^i[k+l], \boldsymbol{z}_v^i[k+l]]^{\mathrm{T}}\in\mathbb{R}^6$ are the state and the reference state of the $i$\textsuperscript{th} UAV. 
According to \eqref{referencemodel_closeloop}, the dynamics of the linear reference model at $k+l$ are expressed as

\begin{flalign}
	\begin{split}
	\boldsymbol{z}^{i}[k+l+1]&=\boldsymbol{z}^{i}[k+l]+h\boldsymbol{z}_v^i[k+l],\\
	\boldsymbol{z}_v^i[k+l+1]&=\boldsymbol{z}_v^i[k+l]+h\boldsymbol{u}^i_{f}[k+l].
	\label{referencemodel_closeloop_kl}
   \end{split}
   \end{flalign}
Note that predicting the reference-model outputs at future steps
$k+l$ $(l\geq1)$ requires relative-state information at those steps,
which is unavailable at the current time.
To avoid exchanging predicted trajectories among neighboring UAVs,
the reference-model input is assumed to remain constant over the
prediction horizon, i.e.,
$\boldsymbol{u}_f^i[k+l]=\boldsymbol{u}_f^i[k]$.
Since the MPC is implemented in a receding-horizon manner, only the
first control input is applied at each sampling instant, and
$\boldsymbol{u}_f^i[k]$ is recomputed using the latest neighboring
information at the next instant.
Therefore, the formation reference can be continuously updated without
exchanging horizon-wise predicted trajectories, thereby reducing the
inter-UAV communication burden.
   
$2)$ \emph{Control Effort Penalty}:
\begin{equation}
\ell_{u}^i=\left\|\hat{\boldsymbol{u}}^i[l|k]\right\|^2_{\boldsymbol{W}_u^i},
\end{equation}
where $\boldsymbol{W}_u^i\in\mathbb{R}^{3\times 3}$ is the weighting matrix,
$\hat{\boldsymbol{u}}^i[l|k]\in\mathbb{R}^3$ is the control input of the $i$\textsuperscript{th} UAV.

Next, we formulate the constraints for the MPC.

 $1)$ \emph{Obstacle-Avoidance Constraints}:
Static obstacles detected within the onboard perception range are
indexed by $m\in\mathcal{O}=\{1,\ldots, N_{\mathrm{obs}}\}$, where $N_{\mathrm{obs}}$ denotes
the number of detected obstacles. After removing the point cloud returns
associated with neighboring UAVs, Euclidean clustering \cite{rusu2008towards} is applied to
extract the centroid and dimensions of each obstacle. The resulting
bounding box is further represented by an enclosing ellipsoid for constructing the
obstacle-avoidance constraints:
\begin{equation}
 b_o^m(\hat{\boldsymbol{p}}^i[l|k])=\left\| \boldsymbol{\Theta}^{-1}_{m}(\hat{\boldsymbol{p}}^i[l|k]-\boldsymbol{p}_o^m)\right\|^2_2 -  1\geq 0, m \in \mathcal{O},
 \label{equ:static_collision_avoidance}
\end{equation}
where $\boldsymbol{p}_o^m\in\mathbb{R}^3$ denotes the centroid of the
$m$th obstacle, and $\boldsymbol{\Theta}_{m}\in\mathbb{R}^{3\times3}$
denotes its ellipsoidal shape matrix. Throughout this paper,
$\boldsymbol{\Theta}_{(\cdot)}$ is used to denote the shape matrix of an
ellipsoidal region. For an axis-aligned ellipsoid,
$\boldsymbol{\Theta}_{(\cdot)}
=\mathrm{diag}(\theta_{(\cdot)}^a,\theta_{(\cdot)}^b,\theta_{(\cdot)}^c)$,
where the diagonal entries represent the corresponding semi-axis lengths.

To enhance the robustness of obstacle avoidance against external
disturbances, a tightened control barrier function (CBF) constraint is
developed. Although the feedforward compensation based on disturbance
estimation can effectively attenuate the disturbance effects, the
remaining uncompensated disturbance may still induce prediction errors
and degrade the safety margin. The proposed CBF constraint explicitly accounts for the one-step
position deviation induced by the bounded residual disturbance.

Define the residual disturbance as
$\tilde{\boldsymbol{d}}^i = \boldsymbol{d}^i - \hat{\boldsymbol{d}}^i$,
which is assumed to satisfy
$\|\tilde{\boldsymbol{d}}^i\|_2 \leq \bar d^i$,
where $\bar d^i > 0$ denotes a predefined upper bound.
Assuming that $\tilde{\boldsymbol{d}}^i$ remains constant over a single sampling period $h$, the resulting one-step position deviation under the translational dynamics \eqref{eqn:trans_dynamics} is given by
$\Delta\boldsymbol{p}^i = \frac{1}{2}h^2\tilde{\boldsymbol{d}}^i$.
Consequently, the norm of this one-step position deviation is bounded by
\begin{equation}
\|\Delta\boldsymbol{p}^i\|_2
\leq
\frac{1}{2}h^2\bar d^i .
\label{equ:residual_position_error}
\end{equation}

\begin{figure}[htbp]
\centering
\includegraphics[width=1.0\linewidth]{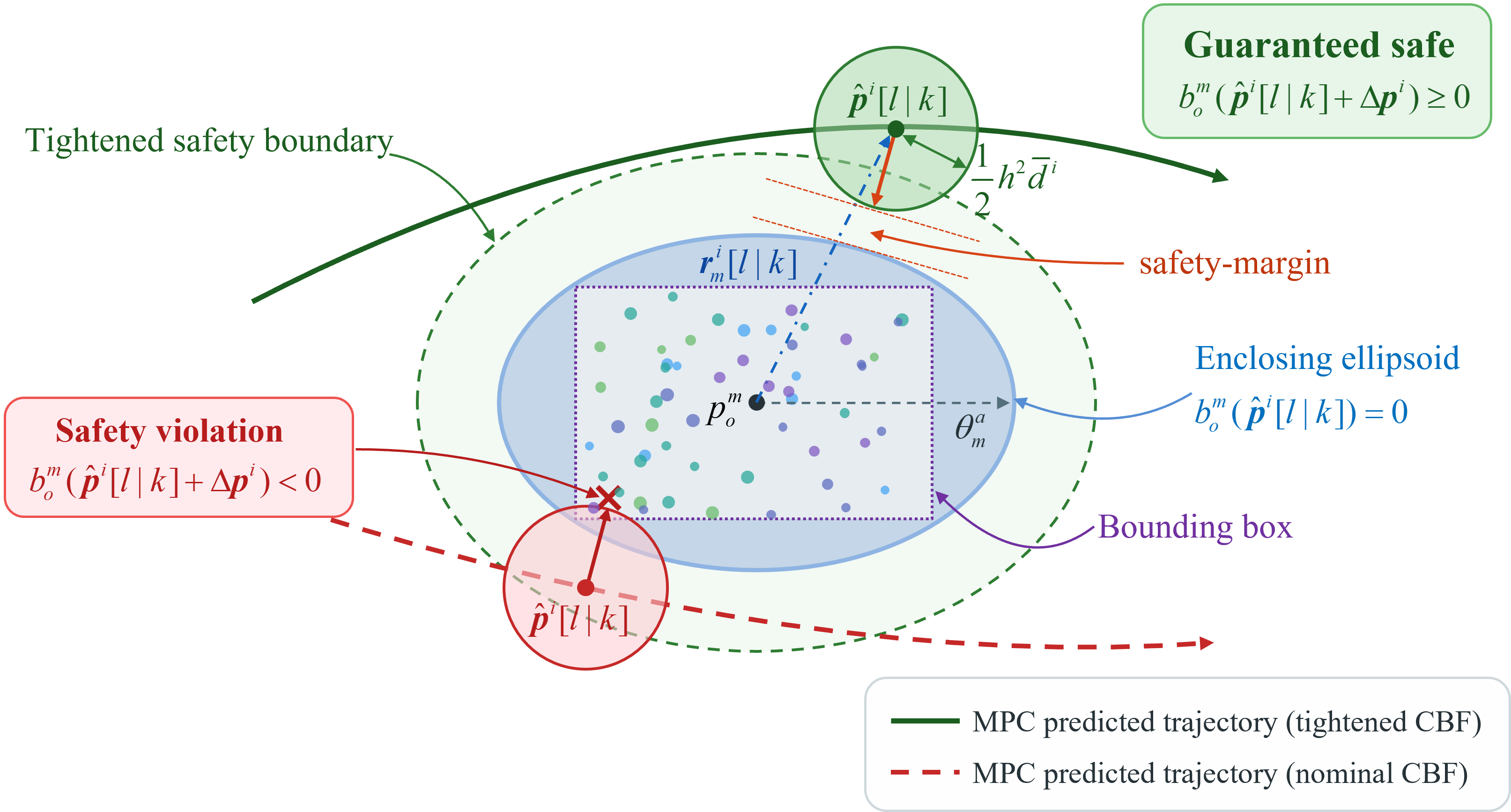}
\caption{ Top-view illustration of the tightened CBF constraint. The colored dots represent the clustered obstacle point cloud. The inner blue ellipsoid denotes the obstacle-avoidance ellipsoid in \eqref{equ:static_collision_avoidance}, while the outer green ellipsoid represents the tightened safety boundary corresponding to \eqref{equ:cbfconstraint}. By reserving a safety margin of radius $\frac{1}{2}h^2\bar{d}^i$, the tightened CBF constraint prevents safety violations along the MPC predicted trajectory under the one-step deviation $\Delta\boldsymbol{p}^i$.  }
\label{fig: cbf-tight}
\end{figure}

For notational convenience, define the symmetric positive-definite matrix
$
\boldsymbol{S}_m
=
\boldsymbol{\Theta}_m^{-\mathrm{T}}
\boldsymbol{\Theta}_m^{-1},
$
where $\boldsymbol{S}_m\succ 0$ characterizes the geometry of the
ellipsoidal obstacle. Then, the corresponding obstacle barrier function
can be equivalently written as
\begin{equation}
b_o^m(\boldsymbol{p})
=
(\boldsymbol{p}-\boldsymbol{p}_o^m)^{\mathrm{T}}
\boldsymbol{S}_m
(\boldsymbol{p}-\boldsymbol{p}_o^m)-1 .
\end{equation}

Let
$
\boldsymbol{r}_m^i[l+1|k]
=
\hat{\boldsymbol{p}}^i[l+1|k]
-
\boldsymbol{p}_o^m .
$
Considering the position uncertainty induced by the residual disturbance,
the barrier function evaluated at the actual position satisfies
\begin{equation}
\begin{aligned}
&b_o^m
(
\hat{\boldsymbol{p}}^i[l+1|k]
+
\Delta\boldsymbol{p}^i
)
\\
=&\,
b_o^m
(
\hat{\boldsymbol{p}}^i[l+1|k]
)
+
2
(\boldsymbol{r}_m^i[l+1|k])^{\mathrm{T}}
\boldsymbol{S}_m
\Delta\boldsymbol{p}^i
\\
&+
(\Delta\boldsymbol{p}^i)^{\mathrm{T}}
\boldsymbol{S}_m
\Delta\boldsymbol{p}^i .
\end{aligned}
\end{equation}

Since $\boldsymbol{S}_m\succ 0$, the quadratic term is non-negative.
Therefore, a conservative lower bound can be obtained by neglecting this
term and applying the Cauchy--Schwarz inequality to the cross term:
\begin{equation}
\begin{aligned}
&b_o^m
(
\hat{\boldsymbol{p}}^i[l+1|k]
+
\Delta\boldsymbol{p}^i
)
\\
\geq&
\,b_o^m
(
\hat{\boldsymbol{p}}^i[l+1|k]
)
-
h^2\bar d^i
\left\| \boldsymbol{S}_m \boldsymbol{r}_m^i[l+1|k] \right\|_2.
\end{aligned}
\label{equ:barrier_lower_bound}
\end{equation}

Substituting the lower bound \eqref{equ:barrier_lower_bound} into the discrete-time CBF condition \eqref{equ:cbf_def} yields the following tightened CBF constraint:
\begin{equation}
\begin{aligned}
&
b_o^m
(
\hat{\boldsymbol{p}}^i[l+1|k]
)
-
(1-\lambda_o)
b_o^m
(
\hat{\boldsymbol{p}}^i[l|k]
)
\\
&\quad
-
h^2\bar d^i
\left\|
\boldsymbol{S}_m
(
\hat{\boldsymbol{p}}^i[l+1|k]
-
\boldsymbol{p}_o^m
)
\right\|_2
\geq0 ,
\end{aligned}
\label{equ:cbfconstraint}
\end{equation}
where $\lambda_o\in(0,1]$, $m\in \mathcal{O}$, and $l=0,\dots,N-1$.  A top-view geometric illustration of the tightened CBF constraint \eqref{equ:cbfconstraint} is shown in Fig.~\ref{fig: cbf-tight}.

$2)$ \emph{Predictive Inter-UAV Collision-Avoidance Constraints}:
The inter-UAV collision-avoidance constraints are designed to prevent UAVs from colliding with each other.
Considering the downwash effect caused by the  high-speed rotation of propellers, a collision region is adopted to describe the
required separation between the UAVs:
 \begin{equation}
	 \boldsymbol{\mathcal{C}}^{\mathrm{int}}=\{\boldsymbol{s}\in\mathbb{R}^3|\left\|{\boldsymbol{\Theta}}_{\mathrm{int}}^{-1}\boldsymbol{s}\right\|_2\leq1\},
 \end{equation}
where ${\boldsymbol{\Theta}}_{\mathrm{int}} = \mathrm{diag}({\theta}^a_{\mathrm{int}},{\theta}^b_{\mathrm{int}},{\theta}^c_{\mathrm{int}})\in\mathbb{R}^{3 \times 3}$ with ${\theta}^a_{\mathrm{int}}={\theta}^b_{\mathrm{int}}<{\theta}^c_{\mathrm{int}}$. 
The larger separation ${\theta}^c_{\mathrm{int}}$ is deliberately set along the vertical direction to mitigate the downwash effect.

Unlike most existing approaches \cite{quan2022distributed, luis2019trajectory},
which require explicit trajectory exchange among UAVs, our method constructs
inter-UAV collision-avoidance constraints using only the state information
received from neighboring UAVs. Specifically, each UAV employs a Kalman
filter based on a constant-velocity motion model to predict the short-term future motion of its neighboring UAVs over a
prediction horizon $N_p\in\mathbb{N}^{+}$, where $N_p<N$. The associated
prediction uncertainty is quantified through covariance propagation and is
subsequently incorporated into the inter-UAV collision-avoidance constraints.

For neighboring UAV $j$, its state is defined as
\begin{equation}
\boldsymbol{x}^j[k]
=
[\boldsymbol{p}^j[k],\boldsymbol{v}^j[k]]^{\mathrm T}
\in\mathbb{R}^{6},
\end{equation}
and UAV $i$ maintains a local estimate
$\hat{\boldsymbol{x}}_i^j[k]$ of this state.
A constant-velocity model is adopted for state propagation:
\begin{equation}
\boldsymbol{x}^j[k+1]
=
\boldsymbol{F}\boldsymbol{x}^j[k]
+
\boldsymbol{w}_i^j[k],
\end{equation}
where
\begin{equation}
\boldsymbol{F}
=
\begin{bmatrix}
\boldsymbol I_3 & h\boldsymbol I_3\\
\boldsymbol 0_3 & \boldsymbol I_3
\end{bmatrix},
\end{equation}
and $\boldsymbol{w}_i^j[k]\in\mathbb{R}^{6}$ denotes the zero-mean Gaussian
process noise with covariance $\boldsymbol{Q}_i^j\in\mathbb{R}^{6\times6}$,
accounting for deviations from the constant-velocity assumption.

At each sampling instant $k$, UAV $i$ receives the position and velocity
information transmitted by neighboring UAV $j$ through the inter-UAV
wireless network. The received neighbor-state observation is denoted by
$\boldsymbol{y}_i^j[k]\in\mathbb{R}^{6}$ and modeled as
\begin{equation}
\boldsymbol{y}_i^j[k]
=
\boldsymbol{H}\boldsymbol{x}^j[k]
+
\boldsymbol{\nu}_i^j[k],
\end{equation}
where $\boldsymbol{H}=\boldsymbol{I}_6$, and
$\boldsymbol{\nu}_i^j[k]\in\mathbb{R}^{6}$ denotes zero-mean observation
noise with covariance $\boldsymbol{R}_i^j$.

Let $\hat{\boldsymbol{x}}_i^{j,-}[k]$ and
$\boldsymbol{P}_i^{j,-}[k]$ denote the prior state estimate and its
associated error covariance at time $k$, respectively, where the
superscript ``$-$'' indicates the estimate before incorporating the
neighbor-state observation received at time $k$.
The posterior state estimate and covariance after the measurement update
are denoted by $\hat{\boldsymbol{x}}_i^j[k]$ and
$\boldsymbol{P}_i^j[k]$, respectively.

Based on the posterior estimate at the previous sampling instant $k-1$, the
Kalman filter first performs the time update as
\begin{equation}
\begin{aligned}
\hat{\boldsymbol{x}}_i^{j,-}[k]
&=
\boldsymbol{F}\hat{\boldsymbol{x}}_i^j[k-1],\\
\boldsymbol{P}_i^{j,-}[k]
&=
\boldsymbol{F}\boldsymbol{P}_i^j[k-1]\boldsymbol{F}^{\mathrm T}
+
\boldsymbol{Q}_i^j.
\label{equ:prior_estimate}
\end{aligned}
\end{equation}

Using the received neighbor-state information, the prior estimate is corrected as
\begin{equation}
\begin{aligned}
\boldsymbol{K}_i^j[k]
&=
\boldsymbol{P}_i^{j,-}[k]\boldsymbol{H}^{\mathrm T}
(
\boldsymbol{H}\boldsymbol{P}_i^{j,-}[k]\boldsymbol{H}^{\mathrm T}
+
\boldsymbol{R}_i^j
)^{-1},\\
\hat{\boldsymbol{x}}_i^j[k]
&=
\hat{\boldsymbol{x}}_i^{j,-}[k]
+
\boldsymbol{K}_i^j[k]
(
\boldsymbol{y}_i^j[k]
-
\boldsymbol{H}\hat{\boldsymbol{x}}_i^{j,-}[k]
),\\
\boldsymbol{P}_i^j[k]
&=
(
\boldsymbol{I}_6-\boldsymbol{K}_i^j[k]\boldsymbol{H}
)
\boldsymbol{P}_i^{j,-}[k].
\end{aligned}
\end{equation}

The resulting posterior estimate $\hat{\boldsymbol{x}}_i^j[k]$ and covariance $\boldsymbol{P}_i^j[k]$ provide the initial
conditions for short-term prediction:
\begin{equation}
\hat{\boldsymbol{x}}_i^j[0|k]
=
\hat{\boldsymbol{x}}_i^j[k],
\qquad
\boldsymbol{P}_i^j[0|k]
=
\boldsymbol{P}_i^j[k].
\end{equation}
The predicted state mean and covariance are recursively propagated as
\begin{align}
\hat{\boldsymbol{x}}_i^j[l+1|k]
&=
\boldsymbol{F}\hat{\boldsymbol{x}}_i^j[l|k],\\
\boldsymbol{P}_i^j[l+1|k]
&=
\boldsymbol{F}\boldsymbol{P}_i^j[l|k]\boldsymbol{F}^{\mathrm T}
+
\boldsymbol{Q}_i^j,
\end{align}
for $l=0,\ldots,N_p-1$.
For subsequent uncertainty characterization, the covariance matrix is
partitioned as
\begin{equation}
\boldsymbol{P}_i^j[l|k]
=
\begin{bmatrix}
\boldsymbol{P}_{pp,i}^j[l|k] &
\boldsymbol{P}_{pv,i}^j[l|k]\\
\boldsymbol{P}_{vp,i}^j[l|k] &
\boldsymbol{P}_{vv,i}^j[l|k]
\end{bmatrix},
\end{equation}
where $\boldsymbol{P}_{pp,i}^j[l|k]\in\mathbb{R}^{3\times3}$ denotes
the position prediction covariance.

The position prediction uncertainty is represented by a $3\sigma$
ellipsoid constructed from the position covariance
$\boldsymbol{P}_{pp,i}^j[l|k]$:
\begin{equation}
\boldsymbol{\mathcal{C}}^{ij}[k+l]
=
\{
\boldsymbol{s}\in\mathbb{R}^3
\mid
\left\|
(\boldsymbol{\Theta}^{ij})^{-1}[k+l]\boldsymbol{s}
\right\|_2
\leq 1
\},
\end{equation}
where the corresponding shape matrix satisfies
\begin{equation}
\boldsymbol{\Theta}^{ij}[k+l]
(\boldsymbol{\Theta}^{ij}[k+l])^{\mathrm T}
=
9\boldsymbol{P}_{pp,i}^j[l|k].
\end{equation}

To account for both the prediction uncertainty and the required
inter-UAV safety separation, the augmented collision-avoidance region
is defined as the Minkowski sum
\begin{equation}
\boldsymbol{\mathcal{C}}_{a}^{ij}[k+l]
=
\boldsymbol{\mathcal{C}}^{ij}[k+l]\oplus\boldsymbol{\mathcal{C}}^{\mathrm{int}}.
\end{equation}

Following \cite{kurzhanski1997ellipsoidal}, an outer
ellipsoid $\boldsymbol{\mathcal{U}}^{ij}[k+l]$ is constructed to conservatively
enclose $\boldsymbol{\mathcal{C}}_{a}^{ij}[k+l]$:
\begin{equation}
\boldsymbol{\mathcal{U}}^{ij}[k+l]
=
\{
\boldsymbol{s}\in\mathbb{R}^3
\mid
\left\|
(\boldsymbol{\Theta}_{\mathrm{out}}^{ij}[k+l])^{-1}
\boldsymbol{s}
\right\|_2
\leq 1
\},
\end{equation}
such that $\boldsymbol{\mathcal{C}}_{a}^{ij}[k+l]
\subseteq
\boldsymbol{\mathcal{U}}^{ij}[k+l]$.
The shape matrix of the outer ellipsoid satisfies
\begin{equation}
\begin{aligned}
\boldsymbol{\Theta}_{\mathrm{out}}^{ij}
(\boldsymbol{\Theta}_{\mathrm{out}}^{ij})^{\mathrm T}
={}&(1+\beta)\boldsymbol{\Theta}^{ij}
(\boldsymbol{\Theta}^{ij})^{\mathrm T} +(1+\frac{1}{\beta})\boldsymbol{\Theta}_{\mathrm{int}}
\boldsymbol{\Theta}_{\mathrm{int}}^{\mathrm T},
\end{aligned}
\label{augmented_model}
\end{equation}
where
$\beta=\frac{\left\|\boldsymbol{\Theta}_{\mathrm{int}}\right\|_F}
{\left\|\boldsymbol{\Theta}_{ij}\right\|_F}$, $\|\cdot\|_F$ denotes the Frobenius norm, and the time index $k+l$ is omitted for notational simplicity.

Based on the augmented collision-avoidance ellipsoid, the inter-UAV collision-avoidance constraint with respect to UAV $j$ is formulated via the safety function $b_r^j(\cdot)$ as
\begin{equation}
\begin{aligned}
 b_r^j(\hat{\boldsymbol{p}}^i[l|k])=\left\| ({\boldsymbol{\Theta}}_{\mathrm{out}}^{ij})^{-1}[k+l](\hat{\boldsymbol{p}}^i[l|k]-\hat{\boldsymbol{p}}_i^j[l|k])\right\|^2_2 -  1\geq 0,
 \end{aligned}
  \label{equ:reavoidance}
\end{equation}
where $l=0,\dots,N_p$.

\begin{figure}[htbp]
\centering
\includegraphics[width=1.0\linewidth]{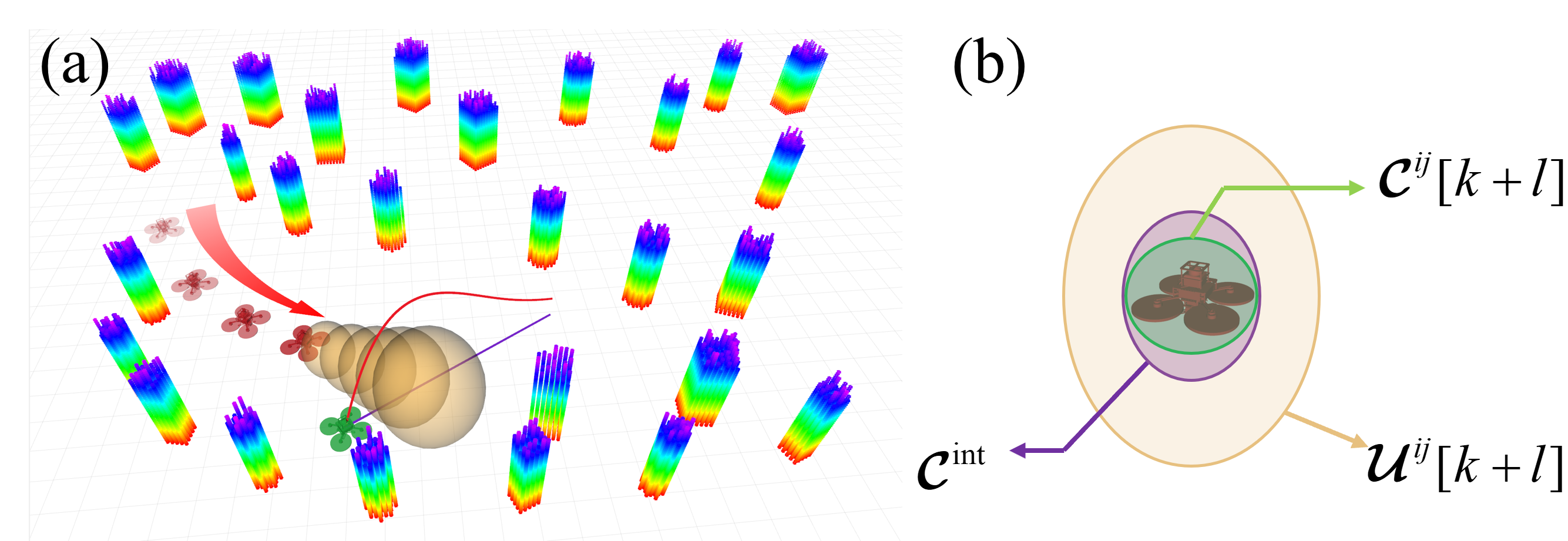}
\caption{Illustration of the predictive inter-UAV collision-avoidance constraints. (a) The lower-priority green UAV actively avoids the red UAV's predicted collision-avoidance ellipsoids $\boldsymbol{\mathcal{U}}^{ij}[k+l]$ (shown in orange), whereas the red UAV does not actively evade. The red and purple lines denote the optimized and formation reference trajectories, respectively. (b) Relationship among the collision region $\boldsymbol{\mathcal{C}}^{\mathrm{int}}$, position prediction uncertainty $\boldsymbol{\mathcal{C}}^{ij}[k+l]$, and outer ellipsoid $\boldsymbol{\mathcal{U}}^{ij}[k+l]$. }
\label{fig: interUAV}
\end{figure}

Meanwhile, a priority mechanism is further adopted to reduce the number of
inter-UAV collision-avoidance constraints in the MPC formulation.
In this mechanism, fixed and distinct priorities are assigned to the UAVs according to their IDs and remain unchanged throughout each mission. Higher-priority UAVs do not need to consider lower-priority UAVs when
constructing collision-avoidance constraints, while lower-priority UAVs must construct collision-avoidance constraints with respect to all higher-priority UAVs. An illustration of the priority-based predictive inter-UAV collision-avoidance constraints is shown in Fig.~\ref{fig: interUAV}.

The benefits of our approach are twofold:
$(1)$ the motion prediction mechanism eliminates the need for exchanging
predicted trajectories among UAVs;
$(2)$ the priority mechanism reduces the number of collision-avoidance constraints included
in the MPC optimization.

$3)$ \emph{Kinodynamic Constraints}:
Several kinodynamic constraints are considered in this paper, including velocity limitations and control input restrictions:
\begin{equation}
        \left\|\hat{\boldsymbol{v}}^i[l|k]\right\|_2 \leq v_{\max}, 
    \left\|\hat{\boldsymbol{u}}^i[l|k]\right\|_2 \leq a_{\max}, \label{Kinodynamicconstrain}
\end{equation}
where $v_{\max}\in \mathbb{R}$ and $a_{\max}\in \mathbb{R}$ denote the maximum velocity and control acceleration, respectively.


Accordingly, the optimal control problem for UAV $i$ is formulated as
\begin{subequations}
\begin{align}
  \boldsymbol{U}_{\mathrm{MPC}}^{i,*} = &\operatorname*{arg\,min}_{\hat{\boldsymbol{U}}^i} \sum_{l=0}^{N-1} \ell(\hat{\boldsymbol{x}}^i[l|k], \hat{\boldsymbol{u}}^i[l|k]), \\
  \text{s.t.} \quad & \text{Dynamics }\eqref{equ:discrete_double_integrator}, \quad l = 0,\dots,N-1, \\
  & \hat{\boldsymbol{x}}^i[0|k] = \boldsymbol{x}^i[k], \\
  & \text{Constraints }\eqref{equ:cbfconstraint}, \eqref{equ:reavoidance}, \text{ and }\eqref{Kinodynamicconstrain},
\end{align}
\label{OCP}
\end{subequations}
\noindent where $\hat{\boldsymbol{U}}^i = \{\hat{\boldsymbol{u}}^i[0|k], \dots, \hat{\boldsymbol{u}}^i[N-1|k]\}$ is the control sequence over the prediction horizon, and the stage cost is given by $\ell(\hat{\boldsymbol{x}}^i[l|k], \hat{\boldsymbol{u}}^i[l|k]) = \ell_{u}^i + \ell_{e}^i$.
By solving \eqref{OCP}, the optimal control sequence $\boldsymbol{U}_{\mathrm{MPC}}^{i,*} = \{\hat{\boldsymbol{u}}^{i,*}[0|k], \dots, \hat{\boldsymbol{u}}^{i,*}[N-1|k]\}$ is obtained. Following the receding-horizon strategy, only the first element, denoted as $\boldsymbol{u}_{\mathrm{MPC}}^{i} = \hat{\boldsymbol{u}}^{i,*}[0|k]$, is applied.

\subsection{Nonlinear-Dynamics-Based Control Realization} \label{nonliner_subsection}
 The
desired control acceleration, incorporating the MPC output and
disturbance feedforward compensation, is given by
\begin{equation}
\boldsymbol{a}_{\mathrm{des}}^i
=
\boldsymbol{u}_{\mathrm{MPC}}^i
-
\hat{\boldsymbol{d}}^i.
\end{equation}
The corresponding
thrust command is then computed as
\begin{equation}
T_{\mathrm{cmd}}^i
=
m^i(\boldsymbol{a}_{\mathrm{des}}^i+\boldsymbol{g})^{\mathrm T}
\boldsymbol{z}_B^i,
\end{equation}
where $\boldsymbol z_B^i=\boldsymbol R(\boldsymbol q^i)\boldsymbol e_3$
denotes the body-frame $z$-axis of UAV $i$ expressed in the inertial frame,
with $\boldsymbol e_3=[0,0,1]^{\mathrm T}$.
The desired attitude is determined from the direction of
$\boldsymbol{a}_{\mathrm{des}}^i+\boldsymbol{g}$.
Meanwhile, following an attitude controller proposed in \cite{faessler2015automatic}, the body-rate command
$\boldsymbol{\omega}_{\mathrm{cmd}}^i$ is generated from the attitude
error. The resulting thrust and body-rate commands
$[T_{\mathrm{cmd}}^i,\boldsymbol{\omega}_{\mathrm{cmd}}^i]$ are then
sent to the PX4 rate controller for execution.

\subsection{Convergence Analysis}\label{subsec:Convergence Analysis}

We now analyze the convergence of the formation dynamics. Specifically, this analysis explicitly accounts for the tracking errors induced by the local MPC. With the tracking errors defined in \eqref{trackingerr}, substituting the reference-model input \eqref{formation_second_order} into the discrete-time double-integrator dynamics \eqref{referencemodel_closeloop} yields
\begin{flalign}
  \begin{aligned}
\boldsymbol{z}^{i}[k+1]&=\boldsymbol{z}^{i}[k]+h\boldsymbol{z}_v^i[k],\\
\boldsymbol{z}^i_v[k+1]&=\boldsymbol{z}^i_v[k]
    - \sum_{j=1}^n a_{i j}[k]\left(\boldsymbol{z}^i[k]-\boldsymbol{z}^j[k]-\boldsymbol{\delta}^{i j}\right)h\\
	&-\sum_{j=1}^n a_{i j}[k]\left(\boldsymbol{e}^i[k]-\boldsymbol{e}^j[k]\right)h
	-2{\gamma}\boldsymbol{e}_v^i[k]h\\
    &-2{\gamma}\left(\boldsymbol{z}_v^i[k]-\boldsymbol{v}^*\right)h.
    \end{aligned}
    \label{formation_second_order_err}
\end{flalign}
To facilitate the formation control analysis, we define
\begin{flalign}
  \begin{aligned}
		\tilde{\boldsymbol{z}}^i[k]&=\boldsymbol{z}^i[k]-\boldsymbol{\delta}^i - kh\boldsymbol{v}^*, \quad \tilde{\boldsymbol{z}}_v^i[k]=\boldsymbol{z}_v^i[k]-\boldsymbol{v}^*, \\
		\tilde{\boldsymbol{y}}^i[k]&=\tilde{\boldsymbol{z}}^i[k]+\left(1 / \gamma\right) \tilde{\boldsymbol{z}}_v^i[k].
    \end{aligned}
    \label{transform_state}
\end{flalign}
We then stack the above transformed states and the tracking errors of all UAVs into the following vectors:
\begin{flalign}
  \begin{aligned}
		\boldsymbol{\eta}[k]&=\left[\tilde{\boldsymbol{z}}^1[k], \tilde{\boldsymbol{y}}^1[k],\ldots, \tilde{\boldsymbol{z}}^n[k], \tilde{\boldsymbol{y}}^n[k]\right]^{\mathrm{T}}\in\mathbb{R}^{6n},\\
		\boldsymbol{\vartheta }[k]&=\left[\boldsymbol{e}^1[k], \boldsymbol{e}_v^1[k],\ldots, \boldsymbol{e}^n[k], \boldsymbol{e}_v^n[k]\right]^{\mathrm{T}}\in\mathbb{R}^{6n}.
    \end{aligned}
    \label{stack_vector}
\end{flalign}
Using the above coordinate transformation, \eqref{formation_second_order_err} can be written in the following compact form:
\begin{flalign}
		\begin{split}
			\boldsymbol{\eta}[k+1]&=\left[\left(\boldsymbol{I}_n \otimes \boldsymbol{A}-\boldsymbol{L}[k] \otimes \boldsymbol{B}\right) \otimes \boldsymbol{I}_3\right] \boldsymbol{\eta}[k]\\
			&+\left[\left(\boldsymbol{I}_n\otimes\boldsymbol{C}-\boldsymbol{L}[k] \otimes \boldsymbol{D}\right)\otimes\boldsymbol{I}_3\right]\boldsymbol{\vartheta}[k] \\
		&\triangleq\left(\boldsymbol{\Xi}[k] \otimes \boldsymbol{I}_3\right) \boldsymbol{\eta}[k]+\left(\boldsymbol{\Psi}[k]\otimes\boldsymbol{I}_3\right)\boldsymbol{\vartheta}[k],
		\label{referencemodel_discrete}
		\end{split}
	\end{flalign}
where $\otimes$ denotes the Kronecker product, $\boldsymbol{L}[k]$ is the Laplacian matrix associated with the interaction graph $\mathcal{G}[k]$, and the matrices $\boldsymbol{A}$, $\boldsymbol{B}$, $\boldsymbol{C}$ and $\boldsymbol{D}$ are expressed as follows
	\begin{equation}
		\begin{aligned}
		\boldsymbol{A}&=\left[\begin{array}{cc}
		-\gamma h+1 & \gamma h \\
		\gamma h & -\gamma h+1
		\end{array}\right], \quad \boldsymbol{B}=\left[\begin{array}{ll}
		0 & 0 \\
		\frac{h}{\gamma} & 0
		\end{array}\right],\\
		\boldsymbol{C}&=\left[\begin{array}{cc}
			0 &  0 \\
			0 & -2h
			\end{array}\right], \quad \boldsymbol{D}=\left[\begin{array}{ll}
			0 & 0 \\
			\frac{h}{\gamma} & 0
			\end{array}\right].
		\end{aligned}
	\end{equation}

We rewrite \eqref{referencemodel_discrete} as the following nonhomogeneous system:
\begin{equation}
\boldsymbol{\eta}[k+1]
=
(\boldsymbol{\Xi}[k]\otimes\boldsymbol{I}_{3})
\boldsymbol{\eta}[k] + \boldsymbol{\chi}[k],
\label{eq:nonhomogeneous_dynamics}
\end{equation}
where
\begin{equation}
\boldsymbol{\chi}[k]
\triangleq
(\boldsymbol{\Psi}[k]\otimes\boldsymbol{I}_{3})
\boldsymbol{\vartheta}[k]
\end{equation}
represents the equivalent perturbation induced by the MPC tracking errors.

The state-transition matrix associated with the homogeneous part of \eqref{eq:nonhomogeneous_dynamics} from time $l$ to $k$ is defined as
\begin{equation}
\boldsymbol{\Phi}(k,l)
=
\boldsymbol{\Xi}[k-1]\cdots
\boldsymbol{\Xi}[l+1]\boldsymbol{\Xi}[l],
\qquad k>l\geq0,
\label{eq:state_transition}
\end{equation}
with $\boldsymbol{\Phi}(k,k)=\boldsymbol{I}_{2n}$ for $k\geq0$.
Based on the state-transition matrix, the solution of the nonhomogeneous system \eqref{eq:nonhomogeneous_dynamics} can be expressed as
\begin{equation}
\begin{aligned}
\boldsymbol{\eta}[k]
=\left(\boldsymbol{\Phi}(k,0)\otimes\boldsymbol{I}_3\right)
\boldsymbol{\eta}[0]
+\sum_{s=0}^{k-1}
\left(\boldsymbol{\Phi}(k,s+1)\otimes\boldsymbol{I}_3\right)
\boldsymbol{\chi}[s].
\label{eq:nonhomogeneous_solution}
\end{aligned}
\end{equation}

It can be observed from \eqref{eq:nonhomogeneous_solution} that both the initial state propagation and the cumulative effect of the perturbation $\boldsymbol{\chi}[s]$ are governed by the state-transition matrix $\boldsymbol{\Phi}(k,l)$. We characterize the convergence of $\boldsymbol{\Phi}(k,l)$ in the following lemma.

\begin{lemma}
Under Assumptions~\ref{assump:finiteset} and \ref{assump:spanningtree}, if the velocity damping gain is selected such that
\begin{equation}
\sqrt{\Delta}\leq\gamma<1/h,
\label{eq:gain_condition}
\end{equation}
where $\Delta=\max\left\{\sum_{j=1,j\neq i}^{n}a_{ij}[k]:1\leq i\leq n,\;k\geq0\right\}$, then $\boldsymbol{\Xi}[k]$ is a row-stochastic matrix for all $k\geq 0$. 
Moreover, for every $l\geq 0$, there exists a nonnegative vector $\boldsymbol{w}_l\in\mathbb{R}^{2n}$ with $\mathbf{1}_{2n}^{\mathrm{T}}\boldsymbol{w}_l=1$, along with constants $\varrho>0$ and $\lambda_f\in(0,1)$ independent of $l$, such that
\begin{equation}
\left\|\boldsymbol{\Phi}(k,l)-\mathbf{1}_{2n}\boldsymbol{w}_l^{\mathrm{T}}\right\|_2 \leq \varrho\lambda_f^{k-l}, \qquad \forall k\geq l\geq 0.
\label{eq:phi_convergence}
\end{equation}
\label{lemma:transition_matrix}
\end{lemma}

\begin{proof}
Under Assumption~\ref{assump:finiteset} and the gain condition \eqref{eq:gain_condition}, it follows directly from Lemma~6 in \cite{qin2012consensus} that $\boldsymbol{\Xi}[k]$ is a row-stochastic matrix with strictly positive diagonal entries and belongs to a finite set. Furthermore, under Assumptions~\ref{assump:finiteset} and \ref{assump:spanningtree}, following the proof of Theorem~3 in \cite{qin2012consensus}, the interval transition matrix $\boldsymbol{\Phi}(k_{r+1},k_r)$ is ergodic, which guarantees that $\lim_{k\rightarrow\infty}\boldsymbol{\Phi}(k,l)=\mathbf{1}_{2n}\boldsymbol{w}_l^{\mathrm{T}}$ with $\boldsymbol{w}_l\in\mathbb{R}^{2n}$, $\boldsymbol{w}_l\geq \mathbf{0}$, and $\mathbf{1}_{2n}^{\mathrm{T}}\boldsymbol{w}_l=1$. Finally, according to the exponential convergence property of products of stochastic matrices \cite{anthonisse1977exponential}, this convergence is exponential, which directly yields \eqref{eq:phi_convergence}.
\end{proof}

\begin{proposition}
Suppose that the tracking errors of the local MPC controllers are uniformly ultimately bounded. Under Assumptions~\ref{assump:finiteset} and \ref{assump:spanningtree}, if the velocity damping gain $\gamma$ satisfies \eqref{eq:gain_condition}, then the formation errors of the UAV swarm are uniformly ultimately bounded.
\label{proposition:uub}
\end{proposition}

\begin{proof}
To establish the convergence of the reference model, we define the common consensus trajectory $\boldsymbol{\varpi}[k]\in\mathbb{R}^3$ as
\begin{equation}
\boldsymbol{\varpi}[k]\triangleq\left(\boldsymbol{w}_0^{\mathrm{T}}\otimes\boldsymbol{I}_3\right)\boldsymbol{\eta}[0]
+\sum_{s=0}^{k-1}\left(\boldsymbol{w}_{s+1}^{\mathrm{T}}\otimes\boldsymbol{I}_3\right)\boldsymbol{\chi}[s].
\end{equation}
Subtracting $\boldsymbol{\varpi}[k]$ from each $3\times 1$ subvector of $\boldsymbol{\eta}[k]$ in \eqref{eq:nonhomogeneous_solution}, the consensus disagreement vector $\tilde{\boldsymbol{\eta}}[k]\triangleq\boldsymbol{\eta}[k]-\left(\mathbf{1}_{2n}\otimes\boldsymbol{I}_3\right)\boldsymbol{\varpi}[k]$ can be expressed as
\begin{equation}
\begin{aligned}
\tilde{\boldsymbol{\eta}}[k]
=&\left[\left(\boldsymbol{\Phi}(k,0)-\mathbf{1}_{2n}\boldsymbol{w}_0^{\mathrm{T}}\right)\otimes\boldsymbol{I}_3\right]\boldsymbol{\eta}[0]\\
&+\sum_{s=0}^{k-1}\left[\left(\boldsymbol{\Phi}(k,s+1)-\mathbf{1}_{2n}\boldsymbol{w}_{s+1}^{\mathrm{T}}\right)\otimes\boldsymbol{I}_3\right]\boldsymbol{\chi}[s].
\end{aligned}
\label{eq:disagreement_dynamics}
\end{equation}
Under Assumption~\ref{assump:finiteset}, $\boldsymbol{\Psi}[k]$ belongs to a finite set. Moreover, since the local MPC tracking errors $\boldsymbol{e}^i[k]$ and $\boldsymbol{e}_v^i[k]$ are uniformly ultimately bounded (UUB), there exists a constant $\bar{\chi}>0$ such that $\|\boldsymbol{\chi}[k]\|_2\leq\bar{\chi}$ for all $k\geq 0$. Taking the Euclidean norm on both sides of \eqref{eq:disagreement_dynamics} and applying Lemma~\ref{lemma:transition_matrix} yield
\begin{equation}
\begin{aligned}
\|\tilde{\boldsymbol{\eta}}[k]\|_2
&\leq \varrho\lambda_f^k\|\boldsymbol{\eta}[0]\|_2
+\sum_{s=0}^{k-1}\varrho\lambda_f^{k-1-s}\bar{\chi}\\
&\leq \varrho\lambda_f^k\|\boldsymbol{\eta}[0]\|_2
+\frac{\varrho\bar{\chi}}{1-\lambda_f}.
\end{aligned}
\label{eq:delta_eta_bound}
\end{equation}
Since $\lambda_f\in(0,1)$, $\varrho\lambda_f^k\|\boldsymbol{\eta}[0]\|_2$ decays exponentially to zero as $k\to\infty$, which implies that $\tilde{\boldsymbol{\eta}}[k]$ is UUB with an ultimate bound of $\frac{\varrho\bar{\chi}}{1-\lambda_f}$.

According to \eqref{stack_vector}, the $(2i-1)$-th and $(2i)$-th $3\times 1$ subvectors of $\tilde{\boldsymbol{\eta}}[k]$ are $\tilde{\boldsymbol{z}}^i[k]-\boldsymbol{\varpi}[k]$ and $\tilde{\boldsymbol{y}}^i[k]-\boldsymbol{\varpi}[k]$, respectively. The reference-model formation errors can be expressed in terms of these subvectors as
\begin{equation}
\begin{aligned}
\boldsymbol{z}^i[k]-\boldsymbol{z}^j[k]-\boldsymbol{\delta}^{ij}
&=\left(\tilde{\boldsymbol{z}}^i[k]-\boldsymbol{\varpi}[k]\right)-\left(\tilde{\boldsymbol{z}}^j[k]-\boldsymbol{\varpi}[k]\right),\\
\boldsymbol{z}_v^i[k]-\boldsymbol{v}^*
&=\gamma\left[\left(\tilde{\boldsymbol{y}}^i[k]-\boldsymbol{\varpi}[k]\right)-\left(\tilde{\boldsymbol{z}}^i[k]-\boldsymbol{\varpi}[k]\right)\right].
\end{aligned}
\label{eq:ref_formation_err}
\end{equation}
Consequently, the UUB property of $\tilde{\boldsymbol{\eta}}[k]$ directly guarantees that the reference-model formation errors $\boldsymbol{z}^i[k]-\boldsymbol{z}^j[k]-\boldsymbol{\delta}^{ij}$ and $\boldsymbol{z}_v^i[k]-\boldsymbol{v}^*$ are UUB.

Finally, the actual formation errors satisfy
\begin{equation}
\begin{aligned}
\|\boldsymbol{p}^i[k]-\boldsymbol{p}^j[k]-\boldsymbol{\delta}^{ij}\|_2
&\leq \|\boldsymbol{z}^i[k]-\boldsymbol{z}^j[k]-\boldsymbol{\delta}^{ij}\|_2\\
&\quad +\|\boldsymbol{e}^i[k]\|_2+\|\boldsymbol{e}^j[k]\|_2,\\
\|\boldsymbol{v}^i[k]-\boldsymbol{v}^*\|_2
&\leq \|\boldsymbol{z}_v^i[k]-\boldsymbol{v}^*\|_2
+\|\boldsymbol{e}_v^i[k]\|_2.
\end{aligned}
\label{eq:actual_formation_err_bound}
\end{equation}
It follows from \eqref{eq:actual_formation_err_bound} that the actual formation errors of the UAV swarm are uniformly ultimately bounded.
\end{proof}

\section{Real-world experiments and simulations}\label{section4}
\subsection{Implementation Details}\label{ImplementationDetails}

\begin{figure}[htbp]
    \centering
    \includegraphics[width=1.0\linewidth]{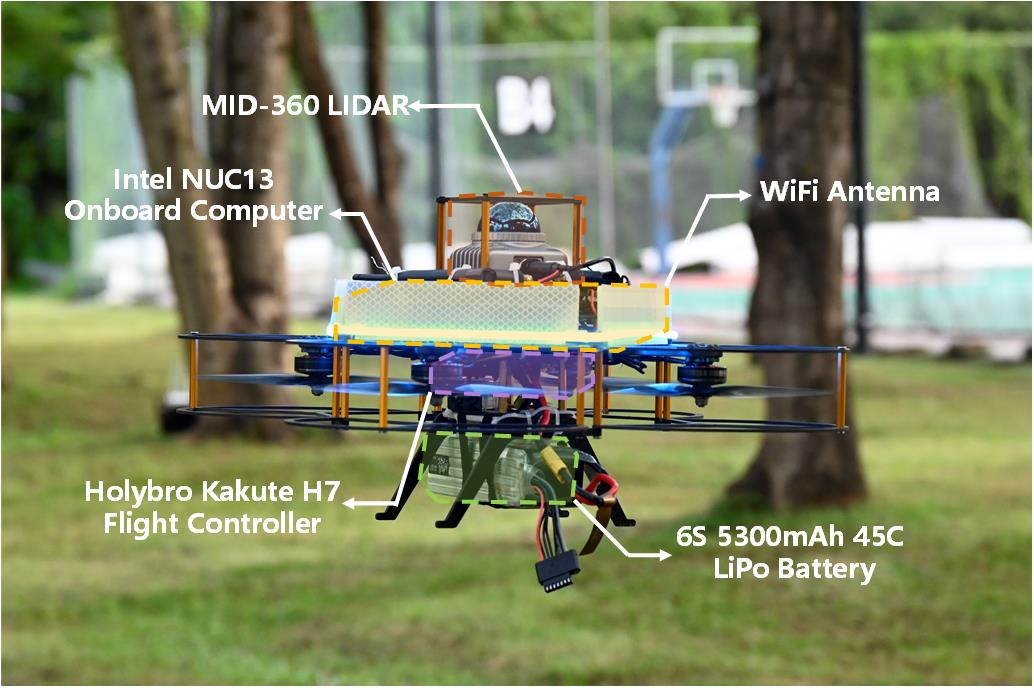}
    \caption{Autonomous UAV platform and onboard hardware components used in the experiments.}
    \label{fig:uav_platform}
\end{figure}

The algorithm is deployed on a fully autonomous and distributed aerial swarm system. Each UAV is equipped with a MID360 LiDAR for localization and obstacle detection, and uses Swarm-LIO2 \cite{zhu2024swarm} to acquire relative position information from neighboring UAVs. The onboard perception module runs at 10\,Hz, continuously detecting obstacles and extracting their position and size.
An overview of the UAV platform is provided in Fig. \ref{fig:uav_platform}.

The local MPC is implemented via the ACADO toolkit \cite{houska2011acado} using a real-time iteration (RTI) scheme with a Gauss--Newton Hessian approximation. It runs at 100\,Hz to ensure real-time execution. All computations are performed onboard using an Intel NUC 13 computer. The controller parameters used in the simulations and experiments are summarized in Table~\ref{table:controller_parameters}.

\begin{table}[h]
\footnotesize
\centering
\caption{Controller Parameters}
\label{table:controller_parameters}
\setlength{\tabcolsep}{4pt}
\begin{tabular}{lclc}
\toprule
Parameter & Value & Parameter & Value \\
\midrule
$\gamma$
& $5$
& $\alpha_1$
& $1$ \\

$\boldsymbol{W}_e^i$
& $\mathrm{diag}[10,10,10,1,1,1]$
& $\alpha_2$
& $1$ \\

$\boldsymbol{W}_u^i$
& $\mathrm{diag}[0.1,0.1,0.1]$
& $\alpha_3$
& $0.12$ \\

$h$
& $0.1\mathrm{s}$
& $\lambda_o$
& $0.5$ \\

$N$
& $10$
& $\bar d^i$
& $0.5 \mathrm{m/s^2}$ \\

$N_p$
& $5$
& &
\\
\bottomrule
\end{tabular}
\end{table}

\subsection{Benchmarks and Ablation Study}\label{Benchmarks}
We evaluate the proposed algorithm through comparative simulations and
ablation studies. First, the proposed collision-avoidance strategy is compared with the DMPC \cite{luis2019trajectory} and BVC \cite{zhou2017fast} algorithms in terms of path length, safety, and communication overhead.  Next, the formation performance is
quantitatively evaluated using a set of formation-error metrics, with
the Swarm Formation algorithm \cite{quan2022distributed} adopted as the
baseline. Subsequently, two ablation studies are conducted under wind disturbances: one evaluates the impact of disturbance-observer-based
compensation, while the other examines the impact of incorporating CBF
constraints into the MPC formulation. 
Finally, a nine-UAV formation flight simulation is conducted in an obstacle-rich environment to evaluate the scalability to a larger swarm size.

\begin{figure}[h]
    \centering
    \includegraphics[width=1.0\linewidth]{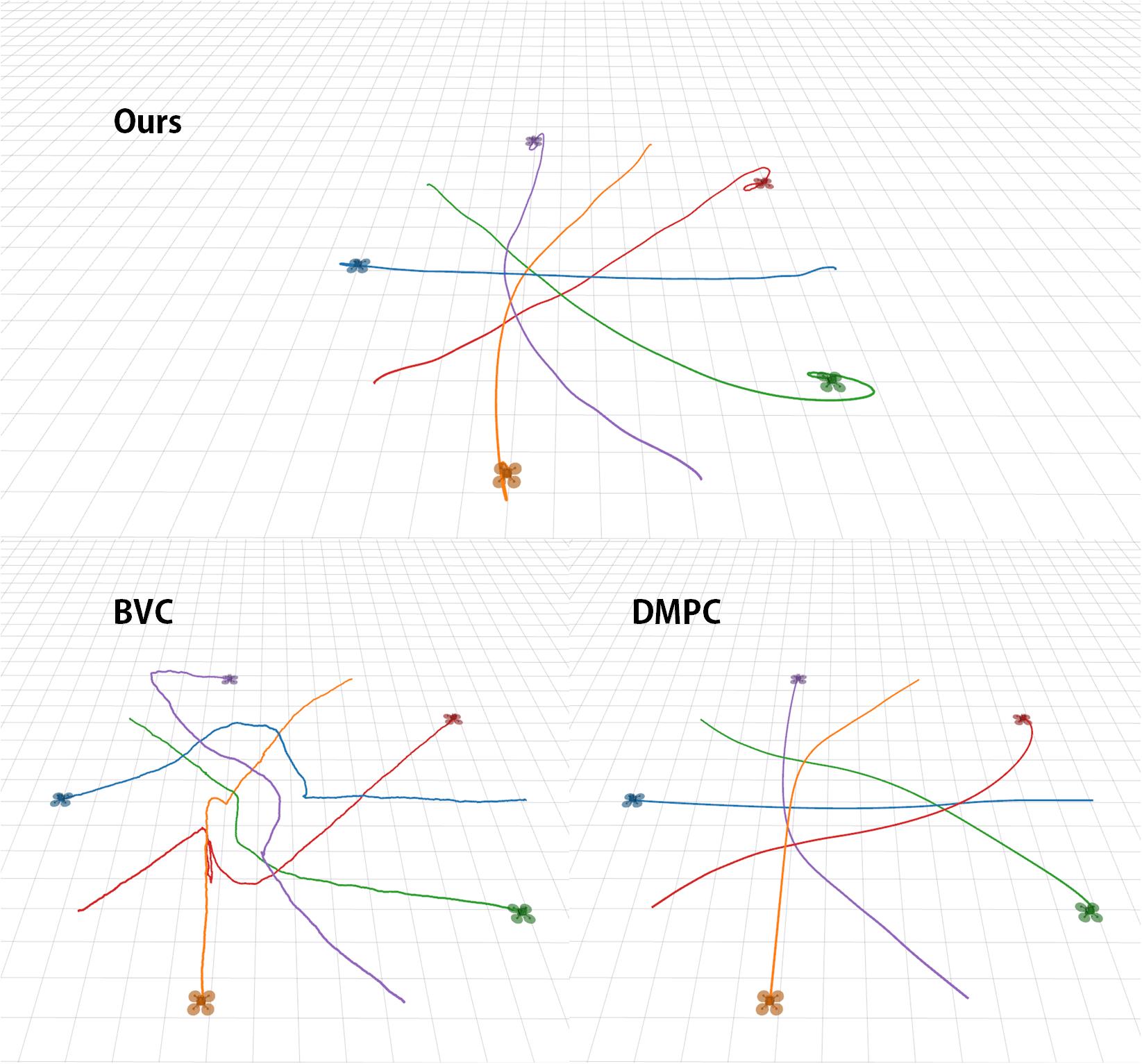}
    \caption{Comparison of collision-avoidance behaviors of DMPC, BVC, and
the proposed method.}
    \label{fig:formation_avoidance}
\end{figure}

\emph{1) Scenario 1. Inter-UAV Collision Avoidance with Five UAVs:}
To evaluate the proposed inter-UAV collision-avoidance strategy, five UAVs are simulated in intersecting flight scenarios, and the proposed method is compared with DMPC and BVC, as shown in Fig.~\ref{fig:formation_avoidance}.
All methods were tested in an open-space environment under identical initial conditions and evaluated through multiple repeated trials to assess the consistency of the results. 
The averaged outcomes are summarized in Table~\ref{table:collision_avoidance}.

\begin{table}[h]
\centering
\footnotesize
\begin{threeparttable}
\caption{Performance Comparison of Inter-UAV Collision-Avoidance Algorithms}
\label{table:collision_avoidance}
\setlength{\tabcolsep}{14pt}
\renewcommand{\arraystretch}{1.1}

\begin{tabular}{l
                S[table-format=2.2]
                S[table-format=1.2]
                S[table-format=1.2]}
\toprule
{\textbf{Method}} &
\multicolumn{1}{c}{\makecell{Avg.\ Len.\\(m)}} &
\multicolumn{1}{c}{\makecell{Min.\ Dist.\\(m)}} &
\multicolumn{1}{c}{\makecell{Comm.\ Load\\(KB/s)}} \\
\midrule
DMPC \cite{luis2019trajectory}          & \bfseries 12.64 & 0.58          & 9.60 \\
BVC \cite{zhou2017fast}           & 17.52           & \bfseries 1.13 & 0.48 \\
\textbf{Ours} & 14.42           & 0.66          & \bfseries 0.96 \\
\bottomrule
\end{tabular}

\begin{tablenotes}\footnotesize
\item \textit{Note:} The communication load is evaluated at 10 Hz with
4-byte floating-point variables. For DMPC, the prediction horizon is
set to $N=10$.
\end{tablenotes}
\end{threeparttable}
\end{table}

To quantitatively evaluate the communication overhead, the communication
load is calculated based on the transmitted data size and communication
frequency. A fully connected communication topology is considered, where
each UAV exchanges information with all other UAVs. All transmitted
variables are represented using single-precision floating-point format,
with 4 bytes required for each variable. For the proposed method, each UAV transmits its current position and velocity to its neighboring UAVs,
resulting in 6 variables per neighbor. In contrast, DMPC requires the
exchange of predicted state trajectories over the prediction horizon. With
$N=10$, each transmitted trajectory contains $6N$ variables. 
BVC exchanges only current position information, corresponding to three variables per neighbor.
The communication load of UAV $i$ is calculated by considering the total
payload exchanged over all associated communication links:
\begin{equation}
\Gamma_i
=
|\mathcal{N}_i| f_c n_d b_d,
\end{equation}
where $|\mathcal{N}_i|$ is the number of neighboring UAVs, $f_c$ denotes
the communication frequency, $n_d$ represents the number of transmitted
variables per neighbor, and $b_d=4$ bytes denotes the data size of each
variable.

As shown by the simulation results, DMPC achieves the shortest path length. BVC achieves a larger minimum separation distance, at the cost of more conservative trajectories, while its
communication overhead is comparable to that of our method. In comparison, the proposed method maintains a moderate path length and sufficient separation while requiring substantially less communication than trajectory-exchange-based DMPC. These results indicate that the
proposed method achieves a favorable balance between trajectory efficiency, inter-UAV collision-avoidance performance, and communication overhead, demonstrating its suitability for distributed
multi-UAV collision-avoidance tasks.


\emph{2) Scenario 2. Formation Generation and Maintenance:}  
The proposed method is compared with the distributed formation planning
method Swarm Formation \cite{quan2022distributed} in terms of formation
generation and formation maintenance.

Considering that the Swarm Formation method aims to achieve formation similarity rather than maintaining a fixed formation geometry, a similarity-based formation error metric is adopted for a fair comparison. Specifically, we formulate a nonlinear optimization problem to find the optimal similarity transformation, including rotation, scaling, and translation, that aligns the desired formation configuration with the current UAV configuration. The similarity-based formation error metric $e_{\text{dist}}$ is defined as
\begin{equation}
e_{\text{dist}} = \min_{\boldsymbol{R},\, \boldsymbol{\tau},\, \kappa} \sum_{i=1}^{n} \left\| \boldsymbol{p}_{c_i} - \left( \kappa\boldsymbol{R} \boldsymbol{p}_{d_i} + \boldsymbol{\tau} \right) \right\|_2.
\label{eq:sim3_error}
\end{equation}
Here, $\boldsymbol{p}_{d_i}$ and $\boldsymbol{p}_{c_i}$ denote the desired formation position and the current position of UAV $i$, respectively. The similarity transformation is composed of a rotation matrix $\boldsymbol{R} \in \mathrm{SO}(3)$, a translation vector $\boldsymbol{\tau} \in \mathbb{R}^3$, and a positive scaling factor $\kappa$. A larger $e_{\text{dist}}$ indicates a greater deviation from the desired shape.

\begin{figure}[h]
    \centering
    \includegraphics[width=1.0\linewidth]{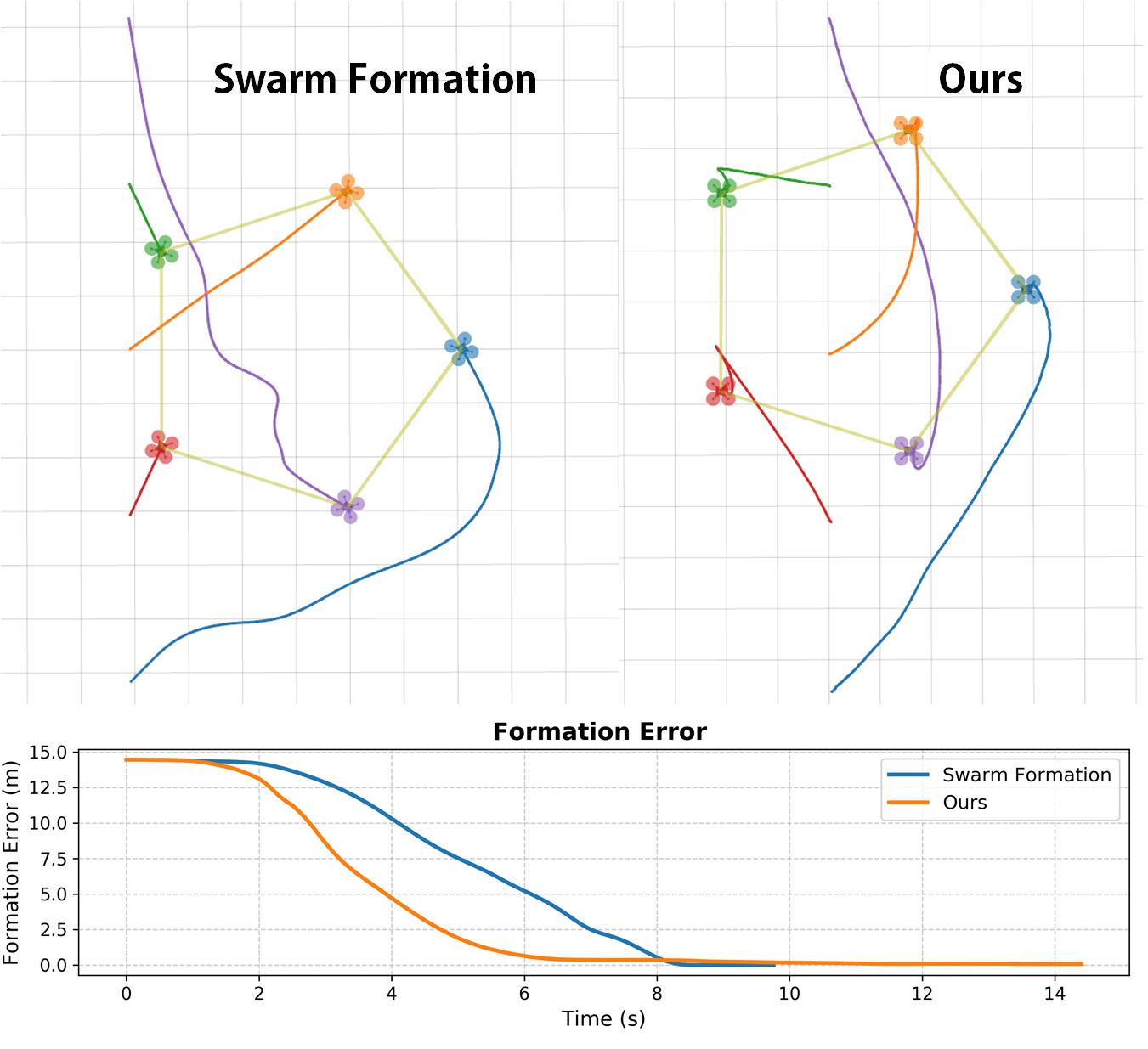}
    \caption{Comparison of formation convergence between Swarm Formation and
the proposed method. Top: UAV trajectories and resulting formations.
Bottom: formation error over time.}
    \label{fig:formation_generation}
\end{figure}

In the formation generation simulation, both methods are initialized with the same UAV configuration and required to achieve the same predefined formation. As shown in Fig. \ref{fig:formation_generation}, both methods successfully converge to the desired formation, while the proposed method requires less time to reach convergence than the Swarm Formation method, indicating improved formation convergence performance.

\begin{figure}[h]
    \centering
    \includegraphics[width=1.0\linewidth]{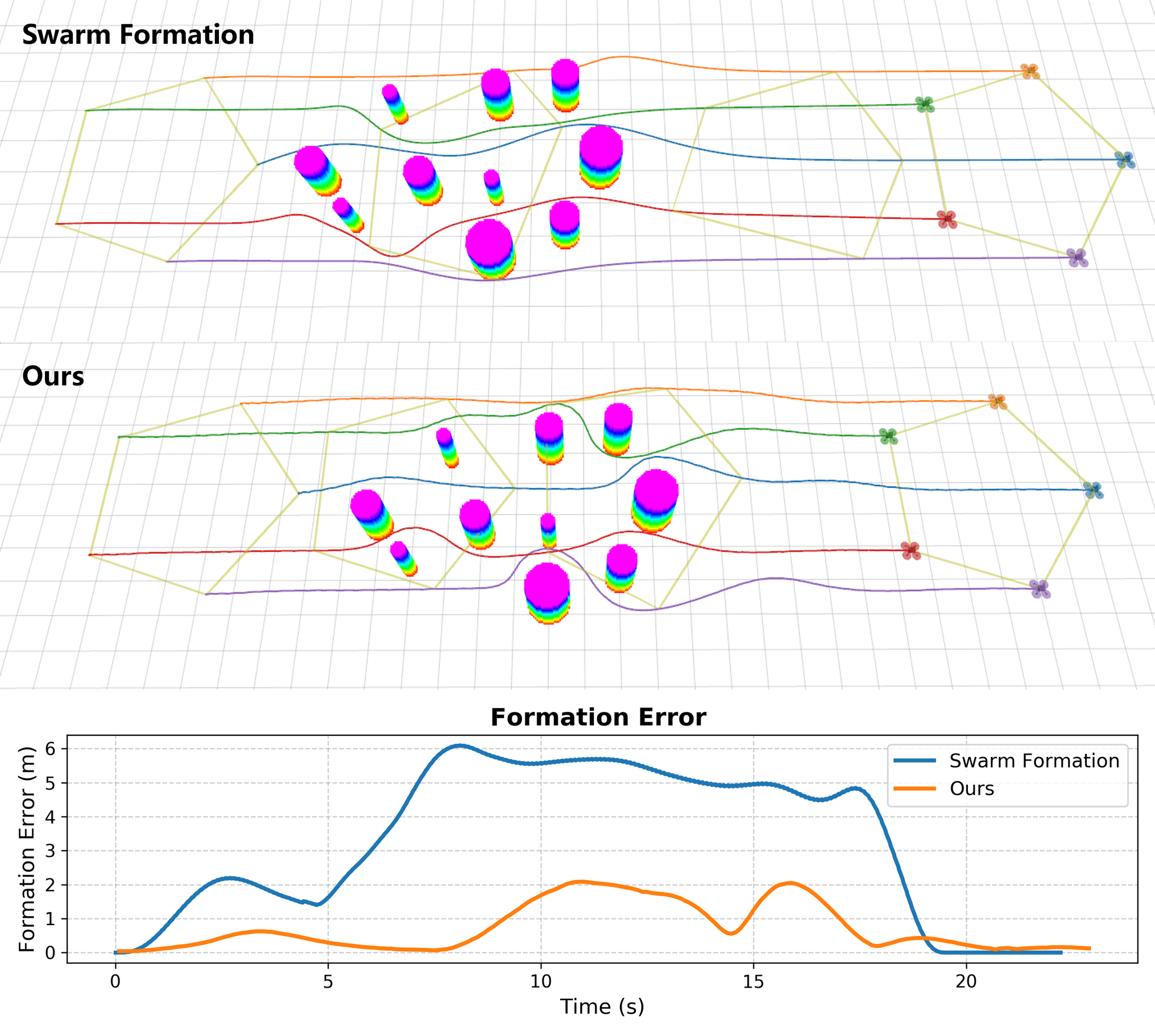}
    \caption{Comparison of formation maintenance performance between Swarm
Formation and the proposed method in an obstacle-rich environment. Top: UAV trajectories in an obstacle-rich environment. Bottom: formation error over time.}
    \label{fig:formation_maintenance}
\end{figure}

In the formation maintenance simulation, all UAVs are initialized in the desired formation and required to traverse multiple obstacle regions. The formation error is recorded at each time step, and its mean value is used to evaluate the formation-maintenance performance. As shown in Fig. \ref{fig:formation_maintenance}, the proposed method yields a smaller mean formation error than the Swarm Formation method, indicating improved formation-maintenance performance during obstacle-rich flight.


\begin{figure}[h]
    \centering
    \includegraphics[width=\linewidth]{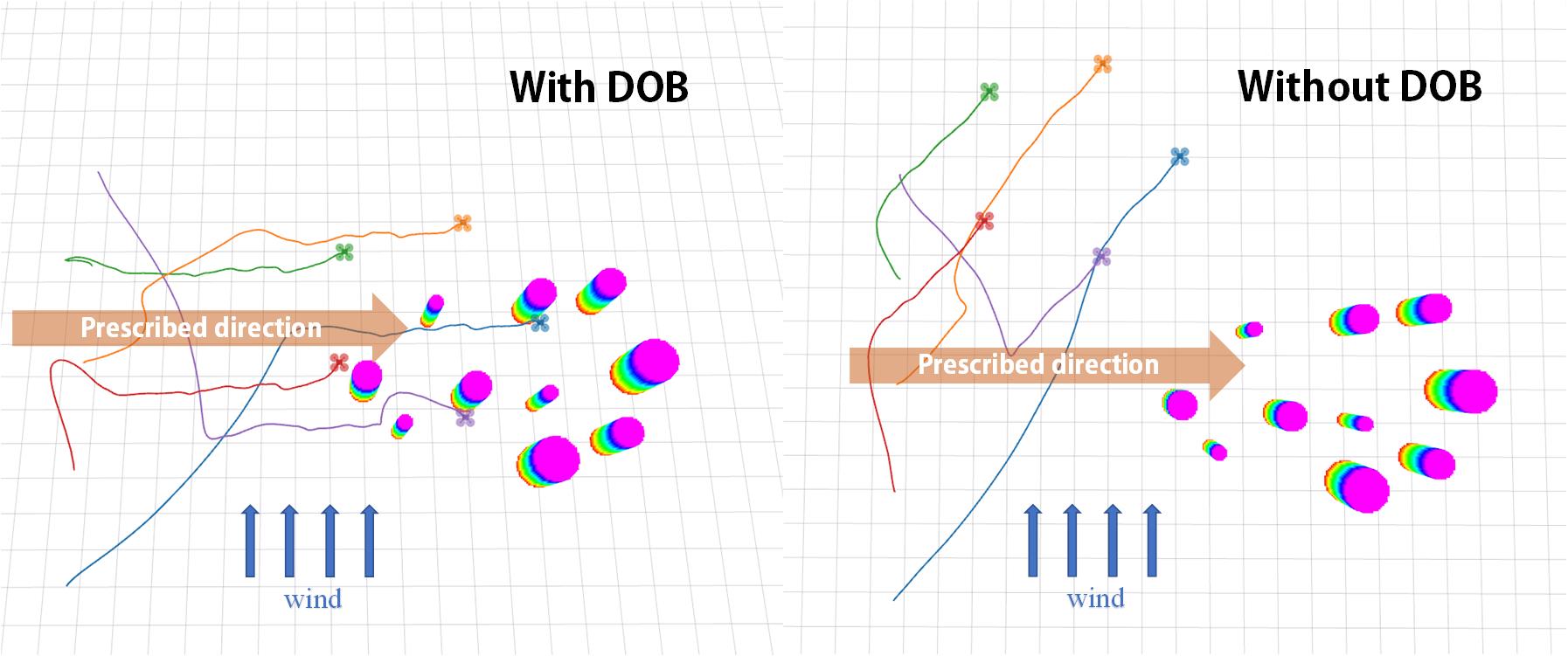}
    \caption{Comparison of formation flight performance with and without the
disturbance observer under wind disturbances. The orange arrow indicates
the desired formation direction, while the blue arrow represents
the wind disturbance direction.}
    \label{fig:Anti-disturbance}
\end{figure}

\emph{3) Scenario 3. Ablation Study on Disturbance Observer under Wind Disturbances:}  
We conduct an ablation simulation to evaluate the effectiveness of the disturbance observer under wind disturbance conditions. The simulation involves five UAVs and is conducted in the Gazebo simulation environment under wind disturbances with an average speed of $3\,\mathrm{m/s}$. The results are shown in Fig. \ref{fig:Anti-disturbance}. 

When disturbance estimation and compensation are disabled, the UAV swarm exhibits a noticeable deviation from the desired flight direction under wind disturbances. In contrast, with the disturbance observer enabled, the external disturbance is estimated online and compensated through a feedforward term combined with the MPC control input, allowing the swarm to better maintain the desired flight direction. These results demonstrate the effectiveness of the disturbance observer in mitigating wind-induced tracking deviations.

\begin{figure}[h]
    \centering
    \includegraphics[width=\linewidth]{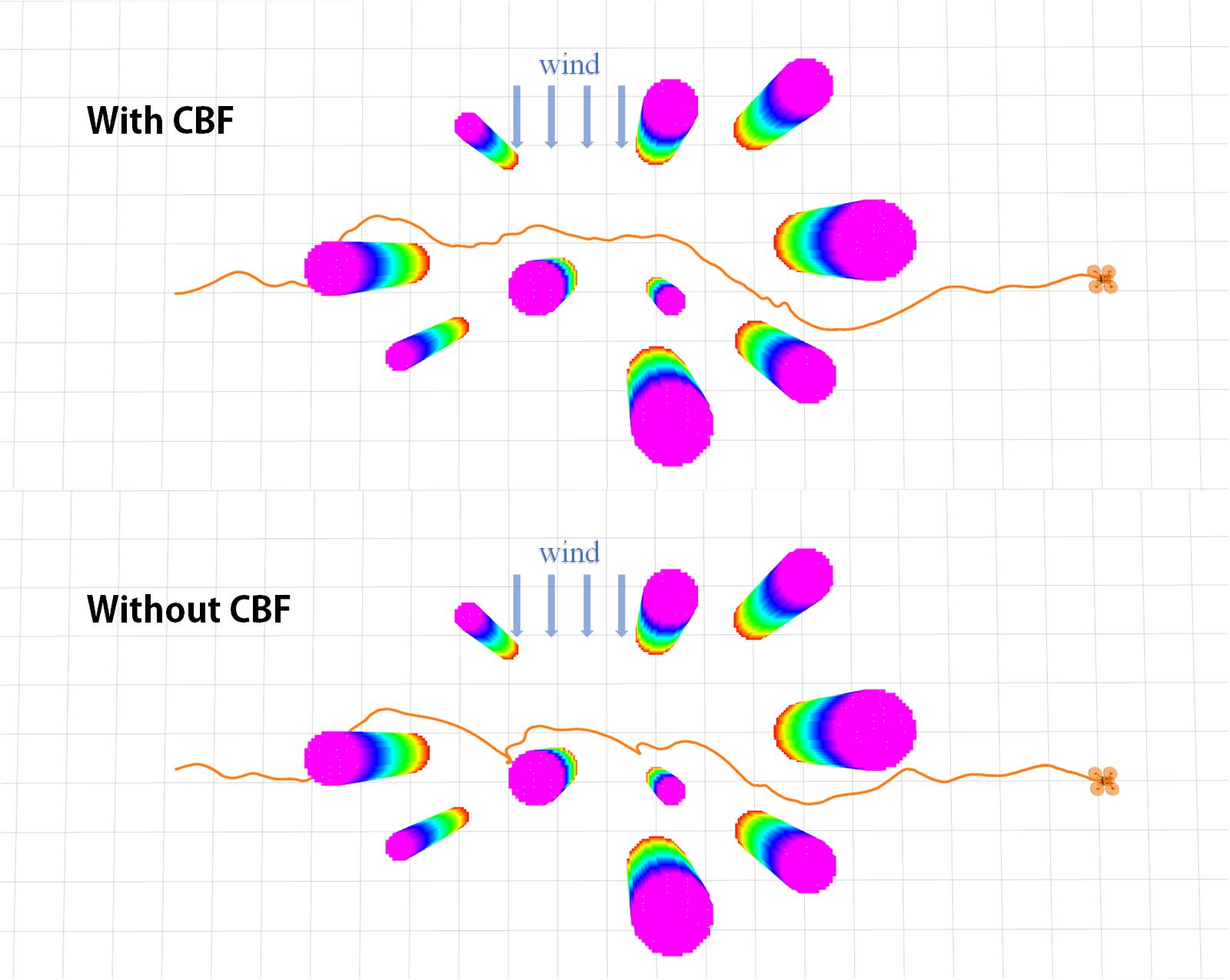}
    \caption{Comparison of collision-avoidance performance with and without CBF-based constraints under wind disturbances. The orange curves represent the UAV flight
trajectories, and the blue arrows indicate the wind direction.}
    \label{fig:cbf}
\end{figure}

\emph{4) Scenario 4. Ablation Study on Control Barrier Function (CBF) under Wind Disturbances:}
Twenty independent trials are conducted to evaluate the effectiveness of the CBF constraint in enhancing flight safety under uncompensated wind disturbances. To isolate the contribution of the CBF, the disturbance observer is completely disabled throughout this simulation. The simulation setup is the same as that in Scenario 3, except that only a single UAV is considered. A representative trial is shown in Fig.~\ref{fig:cbf}.

The performance is evaluated in terms of the minimum safety distance, total trajectory length, and control effort, with the results summarized in Table~\ref{table:cbf_ablation}. For each trial, the minimum distance between the UAV and obstacle surfaces is recorded, and the average value over twenty trials is adopted as the safety metric. The control effort is quantified by the cumulative sum of the UAV acceleration magnitudes. As shown in Table~\ref{table:cbf_ablation}, enabling the CBF constraint increases the minimum safety distance under wind disturbances while maintaining comparable trajectory lengths and control efforts. These results demonstrate that the proposed CBF-based obstacle-avoidance constraint improves flight safety under uncompensated wind disturbances.

\begin{table}[t]
\footnotesize
\centering
\caption{Ablation Results of CBF under Wind Disturbances}
\label{table:cbf_ablation}
\begin{threeparttable}
\setlength{\tabcolsep}{5pt}
\begin{tabular}{lccc}
\toprule
\makecell{Method} 
& \makecell{Avg. Min.\\Dist. (m)} 
& \makecell{Avg. Path\\Len. (m)} 
& \makecell{Avg. Control\\Effort} \\
\midrule
\textbf{With CBF} 
& \textbf{0.69} 
& \textbf{18.10} 
& \textbf{2014.51} \\
Without CBF 
& 0.57 
& 18.89 
& 2025.77 \\
\bottomrule
\end{tabular}
\end{threeparttable}
\end{table}


\emph{5) Scenario 5. Scalability Evaluation with Nine UAVs:}

\begin{figure}[h]
    \centering
    \includegraphics[width=\linewidth]{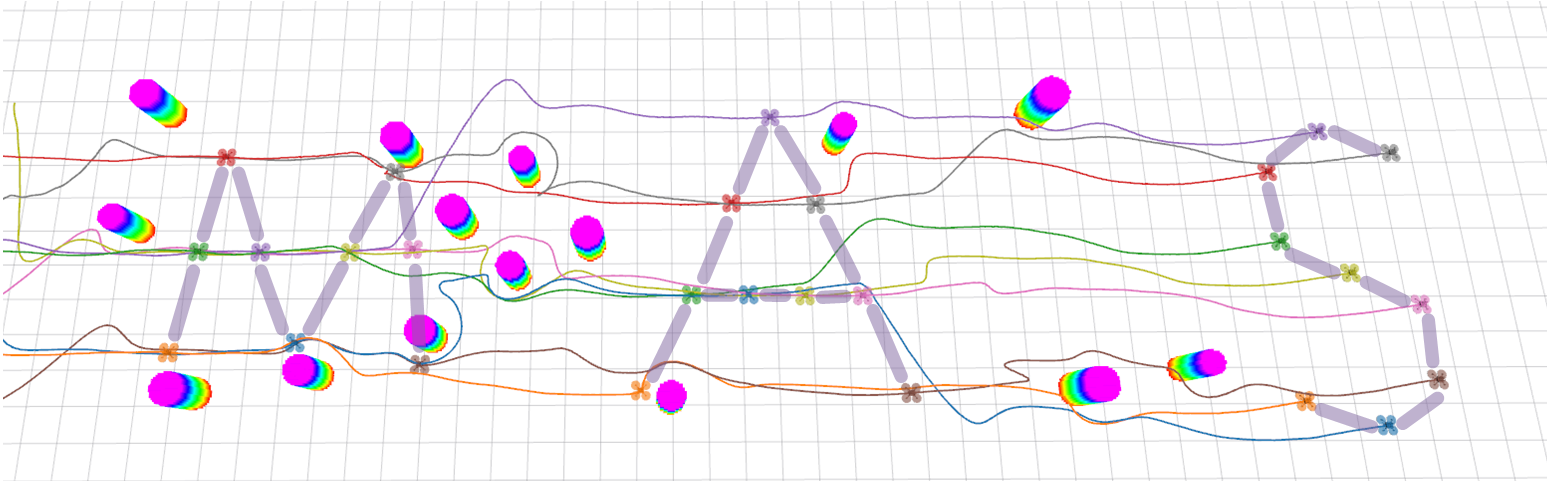}
    \caption{Nine-UAV formation reconfiguration and obstacle avoidance under switching directed communication topologies in a Gazebo simulation environment.}
    \label{fig:nine-drones}
\end{figure}

To evaluate the scalability of the proposed method, a real-time Gazebo
simulation with nine UAVs is conducted on a laptop equipped with an AMD
Ryzen 7 5800H processor. The environment contains randomly distributed
cylindrical obstacles of different sizes.
The UAVs perform successive formation reconfigurations and
communication-topology switching while avoiding obstacles and neighboring
UAVs. Starting from different initial positions, the swarm converges to the
commanded formations and eventually forms the letters ``MAS'', as shown in
Fig.~\ref{fig:nine-drones}. The successful real-time operation of the
nine-UAV swarm provides further evidence of the real-time feasibility of the proposed method for a larger swarm size.

\subsection{Real-World Experiments}\label{RealWorld}

To assess the practical performance of the proposed system, we conduct
real-world experiments covering the following scenarios: formation
with topology switching and formation flight in a forest environment.

Formation convergence is quantitatively evaluated using the following centroid-normalized
formation error:
\begin{equation}
\begin{aligned}
e_{\mathrm{form}}(t)
&=
\frac{1}{n}
\sum_{i=1}^{n}
\left\|
\left(\boldsymbol{p}_i(t)-\bar{\boldsymbol{p}}(t)\right)
-
\left(\boldsymbol{p}_{d_i}(t)-\bar{\boldsymbol{p}}_d(t)\right)
\right\|_2 ,
\end{aligned}
\end{equation}
where $\boldsymbol{p}_i(t)$ and $\boldsymbol{p}_{d_i}(t)$ denote the
actual and desired positions of UAV $i$, respectively.
The variables $\bar{\boldsymbol{p}}(t)$ and
$\bar{\boldsymbol{p}}_d(t)$ represent the actual and desired formation
centroids, respectively. A lower value of $e_{\mathrm{form}}(t)$ indicates better agreement with
the desired formation configuration and improved formation convergence
performance. 

\begin{figure}[h]
    \centering
    \includegraphics[width=\linewidth]{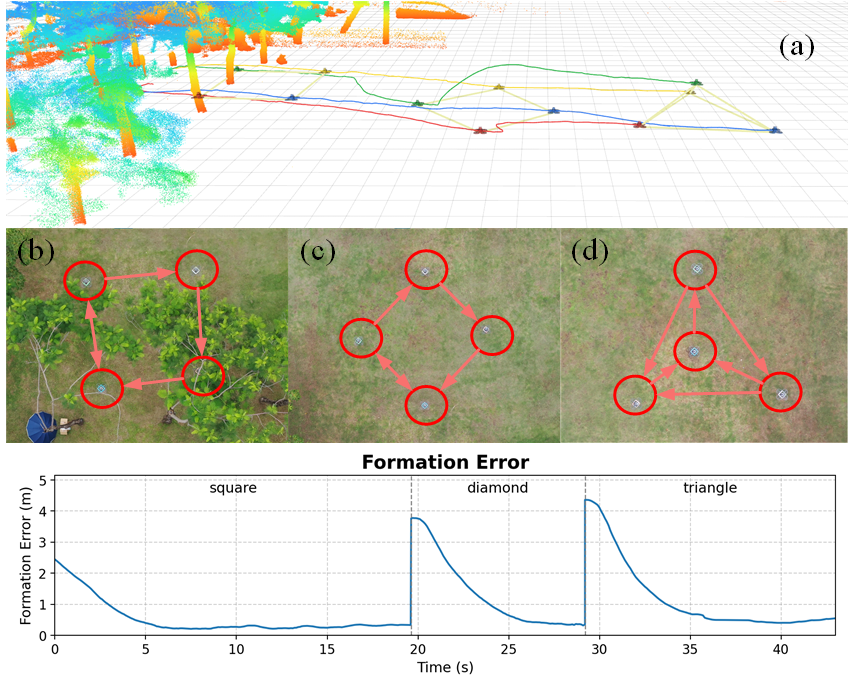}
    \caption{ Outdoor formation flight experiment under switching directed
topologies. (a) shows the flight trajectories during the topology
switching process. (b), (c), and (d) show the UAV formations and corresponding
communication topologies for the square, diamond, and triangular
configurations, respectively, where red arrows indicate the information-flow directions
among UAVs. The bottom plot presents the formation
error during the topology switching process.}
    \label{fig:outdoor_formation_switching}
\end{figure}

\emph{1) Scenario 1. Formation and Topology Switching:}  
To evaluate the effectiveness of the proposed method under switching
topologies, four UAVs are employed to perform formation flight with
time-varying communication topologies. For intuitive visualization, each communication topology is associated with a different formation configuration, with the topology and formation
configuration switched simultaneously. The corresponding flight trajectories, communication
topologies, and formation errors are shown in
Fig.~\ref{fig:outdoor_formation_switching}.
The results show that the UAV swarm converges to each desired formation
after topology switching while maintaining coordinated flight. 
These findings validate the effectiveness of the proposed method in achieving the desired formation under switching directed communication topologies.

\begin{figure}[h]
    \centering
    \includegraphics[width=\linewidth]{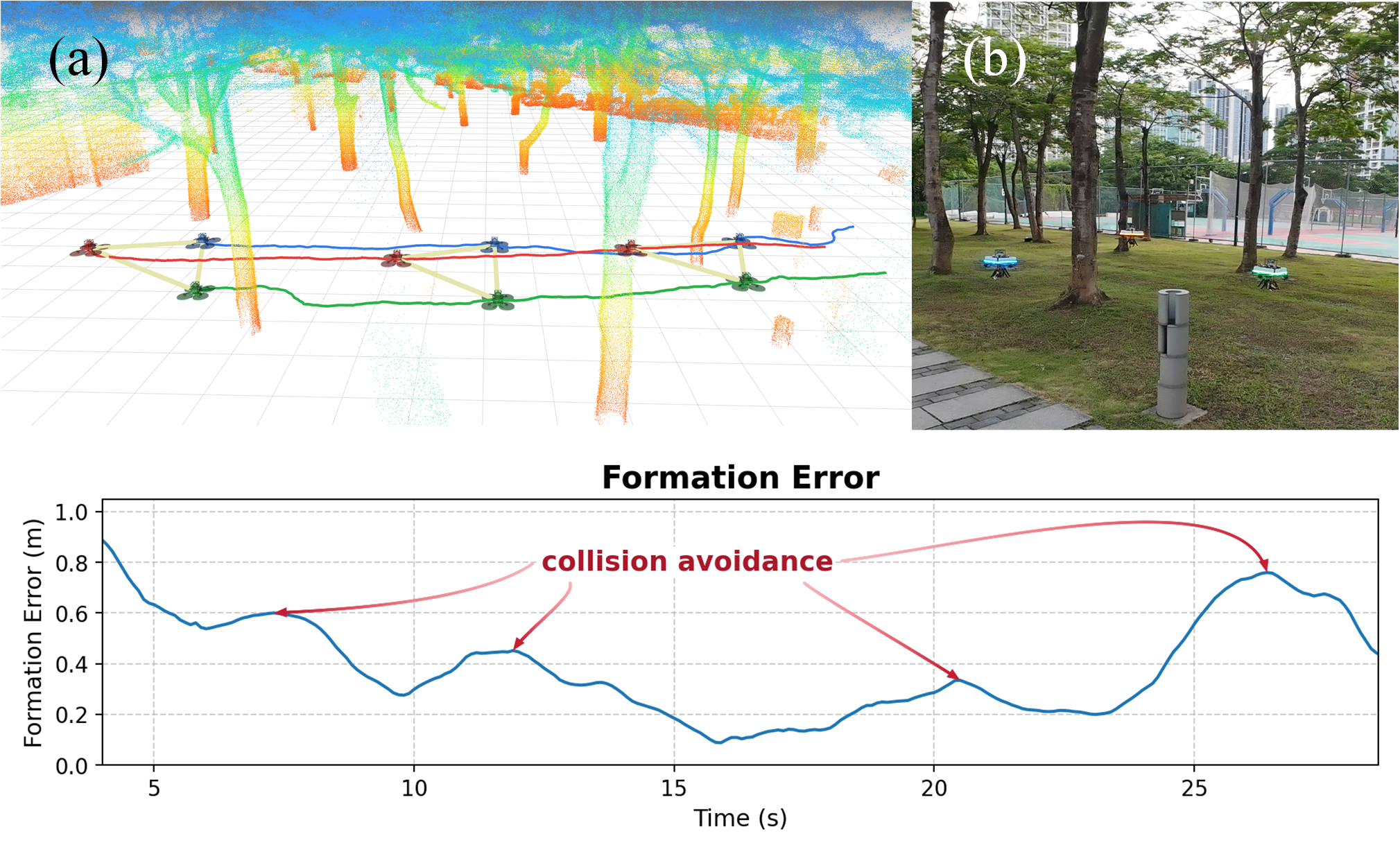}
    \caption{Outdoor experiment of three-UAV formation flying through a forest.
(a) shows the flight trajectories,
while (b) presents a real-world snapshot of the UAV formation during the
flight. The bottom plot
presents the formation error during the flight.}
    \label{fig:outdoor_flying_woods}
\end{figure}

\emph{2) Scenario 2. Three-UAV Forest Flight:}
This experiment evaluates the obstacle avoidance and formation maintenance
performance of the proposed method in a forest environment. Three UAVs
take off from different initial positions and autonomously traverse the
environment while maintaining the desired formation and avoiding
obstacles. As shown in Fig.~\ref{fig:outdoor_flying_woods}, the UAVs
successfully fly through the forest. Due to obstacle-avoidance maneuvers, the swarm formation error remains bounded.
 These results demonstrate the effectiveness of the proposed method in maintaining coordinated formation flight while avoiding obstacles in real-world environments.


\section{Conclusion}\label{section5}

In this paper, we have developed a reference-model-guided distributed
MPC framework for UAV swarm formation under switching directed
topologies. By introducing a local reference model, the proposed method
has decoupled formation coordination from local constrained optimization,
enabling distributed formation control without exchanging predicted
trajectories. A trajectory-exchange-free inter-UAV collision-avoidance strategy has
also been proposed, where neighboring UAV motions have been predicted
locally using a Kalman filter, and the associated prediction uncertainty
has been quantified through covariance propagation. Moreover,
a disturbance-observer-based feedforward compensation scheme and a
CBF-based safety constraint have been integrated to improve robustness and
safety under external disturbances. Extensive comparative simulations,
ablation studies, and flight experiments have demonstrated the
effectiveness of the proposed approach for safe and efficient multi-UAV
formation flight. Future work will focus on integrating the proposed
framework with advanced autonomous perception and decision-making modules
for complex real-world applications.

\section*{Conflict of Interest}
The authors declare that they have no known competing financial interests or personal relationships that could have appeared to influence the work reported in this paper.

\bibliographystyle{IEEEtran}
\bibliography{IEEEabrv,ref}

\newpage

 




\vfill

\end{document}